\documentclass[10pt,journal,compsoc]{IEEEtran}
\usepackage{hyperref}
\usepackage{color}
\usepackage[fleqn]{amsmath}
\usepackage{graphicx}
\usepackage{epigraph}
\usepackage{pifont}
\usepackage{amssymb}
\usepackage{subfig}
\usepackage{rotating}
\usepackage{adjustbox}
\usepackage{booktabs}
\usepackage{multirow}
\usepackage{float}
\usepackage{graphicx}

\usepackage[T1]{fontenc}
\PassOptionsToPackage{normalem}{ulem}
\usepackage{ulem}
\providecommand{\tabularnewline}{\\}

\def\etal{{\it et al. }}

\usepackage{array}
\newcolumntype{L}[1]{>{\raggedright\let\newline\\\arraybackslash\hspace{0pt}}m{#1}}
\newcolumntype{C}[1]{>{\centering\let\newline\\\arraybackslash\hspace{0pt}}m{#1}}
\newcolumntype{R}[1]{>{\raggedleft\let\newline\\\arraybackslash\hspace{0pt}}m{#1}}

\ifCLASSINFOpdf
\else
\fi

\begin{document}

\title{Visual Affordance and Function Understanding: A Survey}

\author{Mohammed Hassanin, Salman Khan, Murat Tahtali
\IEEEcompsocitemizethanks{
\IEEEcompsocthanksitem 
 M. Hassanin and M. Tahtali are with University of New South Wales (UNSW), Canberra, AU. \protect\\
E-mail: ----- 
\IEEEcompsocthanksitem 
S. Khan is with Data61-CSIRO and Australian National University (ANU), Canberra, AU. \protect\\
}
\thanks{Manuscript received --; revised --.}}

\markboth{Journal of \LaTeX\ Class Files,~Vol.~6, No.~1, July~2018}%
{Shell \MakeLowercase{\textit{et al.}}: Bare Demo of IEEEtran.cls for Computer Society Journals}

\IEEEcompsoctitleabstractindextext{
\begin{abstract} 
Nowadays, robots are dominating the manufacturing, entertainment and healthcare industries. Robot vision aims to equip robots with the ability to discover information, understand it and interact with the environment. These capabilities require an agent to effectively understand object affordances and functionalities in complex visual domains. In this literature survey, we first focus on `Visual affordances' and summarize the state of the art as well as open problems and research gaps. Specifically, we discuss sub-problems such as affordance detection, categorization, segmentation and high-level reasoning. Furthermore, we cover functional scene understanding and the prevalent functional descriptors used in the literature. The survey also provides necessary background to the problem, sheds light on its significance and highlights the existing challenges for affordance and functionality learning. 
\end{abstract}

}
\IEEEdisplaynotcompsoctitleabstractindextext
\maketitle



\section{Introduction}
Affordance understanding is concerned with the possible set of actions that an environment allows to an actor. In other words, this area of study aims to answer the question of how an object can be used by an agent? Ecological psychologist James Gibson was the first to introduce the concept of affordances in 1966 \cite{Gibson_Perception_1966}. Since then, the theory of affordances has been extensively used in the design of better and robust robotic systems capable of operating in complex and dynamic environments \cite{horton2012affordances}. In contrast to affordances which are directly dependent on the actor, function understanding relates to identifying the possible set of tasks which can be performed with an object. Object function is therefore a permanent property of an object independent of the characteristics of the user. Affordance and function understanding not only allow humans or AI agents to better interact with the world, but also provide valuable feedback to the product designers who need to consider possible interactions between users and products. As a result, this research topic is highly important for domestic robotics, content analysis and context-aware scene understanding. 

Despite being an indispensable step towards the design of intelligent machines, affordance learning is a complex and highly integrated task. First, to understand how an object can be used by an agent requires reasoning about what it is and where is it located? Furthermore, it is necessary to know about the object's geometry and pose e.g., an inverted cup cannot afford a `\emph{pouring}' action. Unlike traditional classification and detection tasks where each object takes a single label represented as one-hot encoding, a single object can simultaneously take multiple affordances, e.g., a bed is both \emph{`sittable'} and `\emph{layable}'. The object affordances are also dynamic and an intelligent agent should be able to consider both the prior knowledge and past experiences e.g., a cup may be first `\emph{graspable}', then `\emph{liftable}' and finally `\emph{pourable}'. These challenges offer room to novel ideas and innovative solutions for visual affordance learning and functionality understanding.

\begin{figure*} 
    \centering
  \subfloat[]{%
       \includegraphics[clip=true, trim= {1cm 2cm 1cm 0cm}, width=0.56\textwidth]{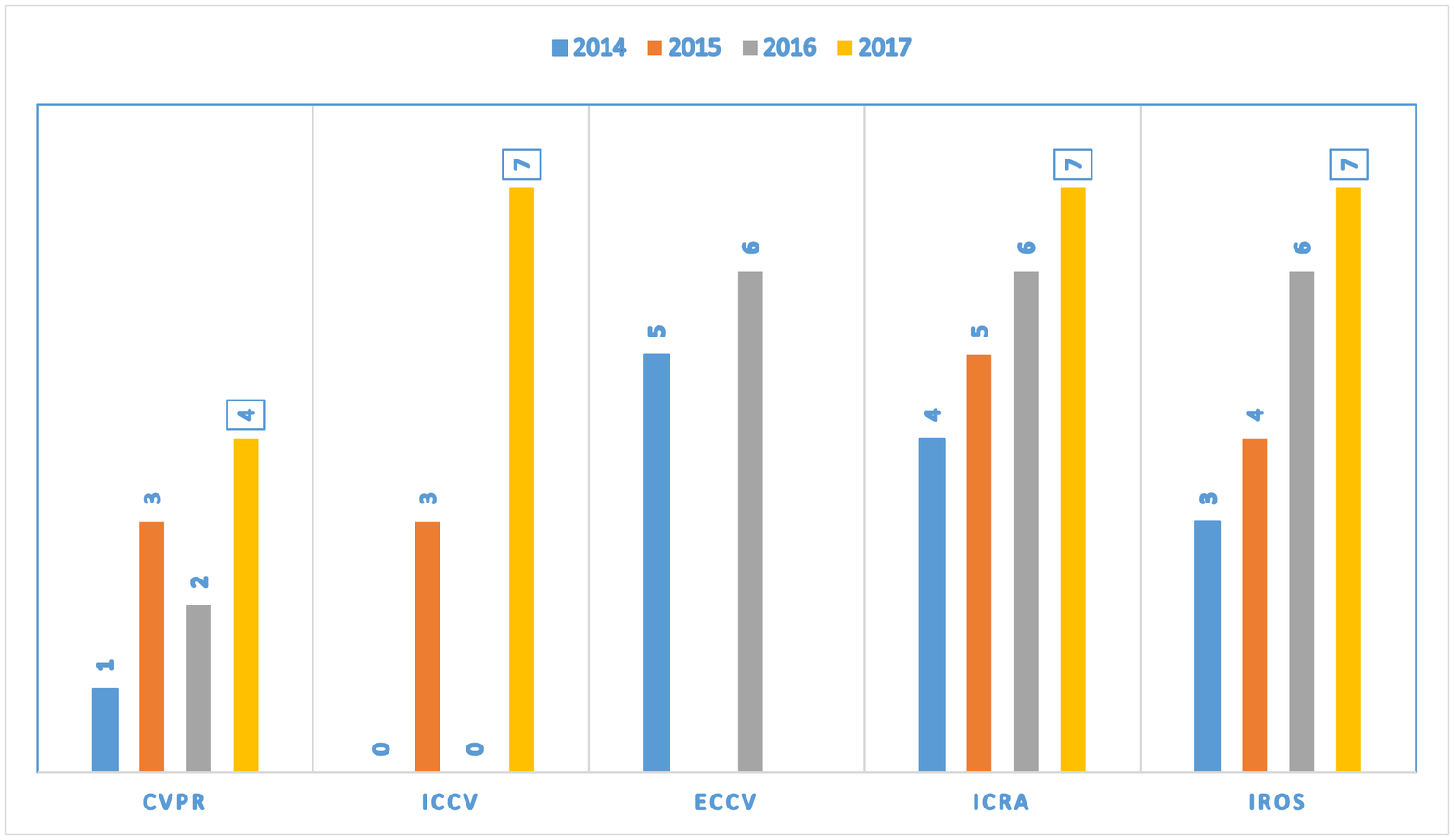}}
     \subfloat[]{%
        \includegraphics[clip=true, trim={0.5cm 1cm 0.5cm 1cm}, width=0.44\textwidth, height=0.35\textwidth]{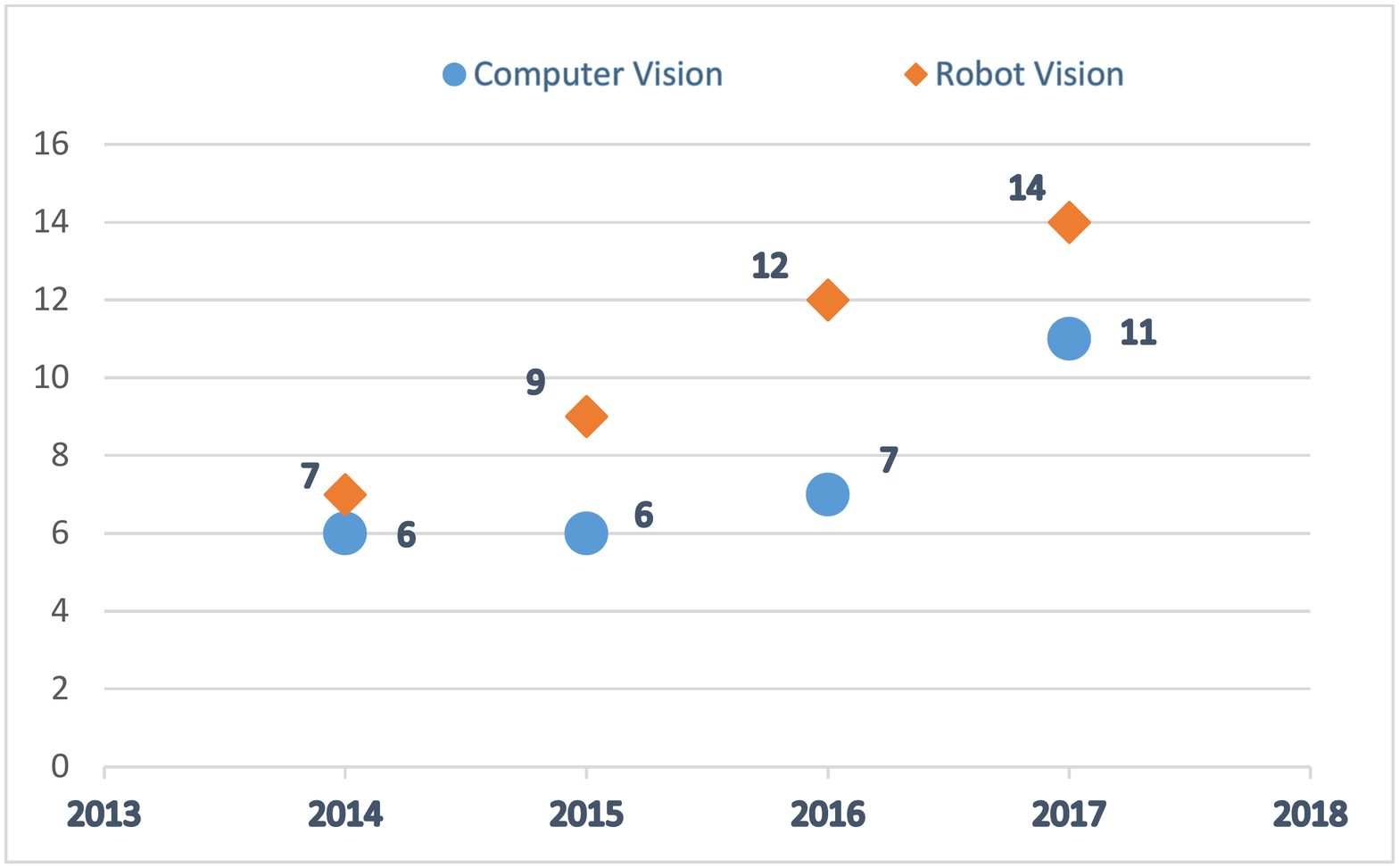}}

  \caption{Growth in the number of papers on visual affordance in computer and robot vision literature in the recent years (from 2014 to 2017).}
  \label{progress} 
\end{figure*}

Affordance learning has been reviewed from different perspectives in the literature;  Bohg et al. \cite{Grasping_survey_2014} reviewed data-driven grasping tasks particularly for manipulation of objects and grasping, Yamanobe et al. \cite{Review_Manipulation_Robots_2017} reviewed the affordance tasks to cover the grasping and manipulation of objects, Min et al. \cite{Survey_Affordance_DevelopmentalRobots_2016} introduced a general overview about the affordances and existing techniques; however, it is devoted to developmental robots and related tasks such as formalization of affordances. 
Up to the best of our knowledge, this survey is the first effort to review the literature from the perspective of \emph{`visual'} affordance and functionality understanding. Notably, other literature reviews cover affordance learning from the aspect of robotics perception, sensory-motor coordination or psychology and neuroscience \cite{jamone2016affordances, Survey_Affordance_DevelopmentalRobots_2016,montesano2008learning, Review_Manipulation_Robots_2017, Grasping_survey_2014}. However, affordance based reasoning is equally important for machine vision and visual scene understanding, as demonstrated by a growing activity in this area (see Figure \ref{progress}).

\begin{figure*}
\centering \includegraphics[width=0.8\textwidth]{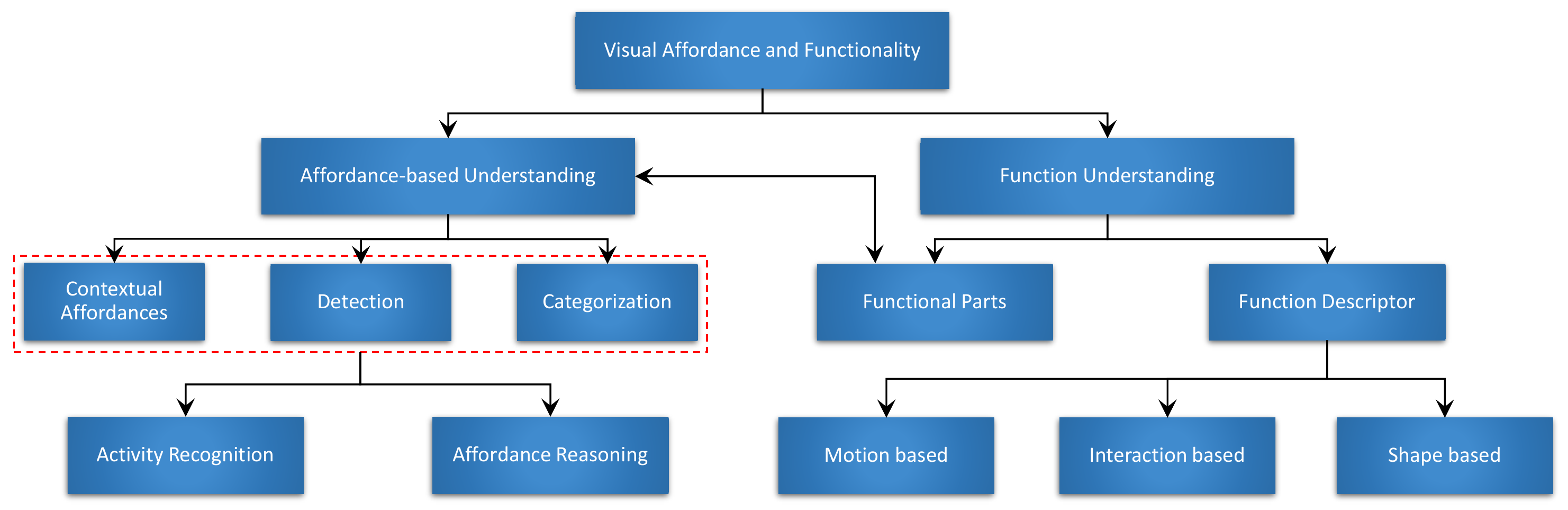}
\caption{Survey taxonomy shows the structure of methods which have been used to solve affordance issues.}\label{fig:taxonomi}
\end{figure*}

Figure \ref{fig:taxonomi} shows the taxonomy of this survey and our scope which we are going to introduce in the following lines. It has two main parts: affordance-based techniques and functionality understanding methods.  The rest of this survey is organized as follows: First, a comprehensive background to the area is provided along with the definition of specific terms frequently used in the visual affordance literature in Sec.~\ref{Terminology and Background}. Afterwards, we provide the significance and challenges in Secs.~\ref{Significance} and \ref{Challenges} respectively. We then cover the research in visual affordance learning in Sec.~\ref{Visual Affordance} and categorize the methods according to specific sub-problems such as affordance detection, categorization, semantic labeling. Efforts to understand functions of different objects and tools are summarized in Sec.\ref{Functional SU}. The main computer vision datasets with affordance and function annotations are listed in Sec.~\ref{Datasets}. Finally, we summarize open research problems in this area and mention new research directions in Sec.~\ref{Open Problems}. 

\section{Terminology and Background}\label{Terminology and Background}
The term affordance was coined by the psychologist Gibson  to define the interactions between an actor and its environment \cite{Gibson_Perception_1966, GibsonAffordanceWord_1977}. In his own words: 
\begin{quotation}
\noindent
\textbf{Affordance: }\emph{``The affordances of the environment are what it offers the animal, what it provides or furnishes, either for good or ill. The word affordance implies the complementarity of the animal and the environment.''}
\end{quotation}
\hfill $-$Gibson, 1979\\
He advocated that the target of computer vision should be to estimate possible interactions between a human, animal and environment in a scene rather than merely detecting the contents of a scene \cite{AffordanceIsTargetofCV_1979}. It was claimed at the time that perception is only a property of agent, but Gibson argued that the perception's meaning is inherited from the environment \cite{Gibson_Perception_1966}. Meanwhile, many researchers tried to find out the best definition of affordance, such as Turvey \cite{Turvey_Definition_Affordance_1992} who defined it as: ``An affordance is a particular
kind of disposition, one whose complement is a dispositional property of an organism". He used the term `effectivity' to denote the dispositional property of an organism (or an AI agent).  Based on the agent's effectivity and the environment affordance properties, an action is realized. Stoffregen \cite{Stoffregen_Affordances_and_Event_2000} sought to clarify the relation between the affordances and actions and concluded that the affordance and actions are not identical in many aspects. In another study of Stoffregen, he critiqued the definition of Turvey that it neglected the relationship between animal and environment, therefore it is sub-optimal from the view of perception-action affordance \cite{Stoffregen_Definition_animal_environment_2003}. Instead of agreeing with their predecessors on the definition of affordance, Sahin et al. \cite{Formalism_Definition_Sahin_2007} introduced a new formalization for affordances. They argued that the affordance has three main constituents: the agent, the environment and the observer. This concept allowed affordances to cover every aspect of robot control ranging from perception to planning i.e., first perceiving the object, then applying behavior to that object and finally the generation of desired effect.

Next, we define several key terms frequently used in the affordance learning literature, that we will consider for the purpose of this survey:\\
\noindent
$-$\textbf{Visual Affordance:}
The term visual affordance means extracting information related to affordance from an image or a video. Similar to other machine vision domains such as object and human activity recognition, it uses computer-vision techniques to perceive the affordance characteristics in visual media.

\begin{figure*} 
    \centering
  \subfloat[Functionality understanding]{%
       \includegraphics[width=0.22\paperwidth,height=0.150\paperheight]{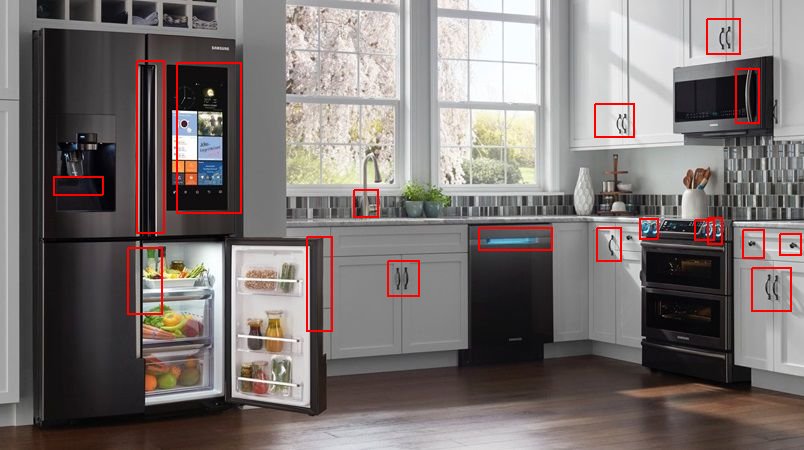}}
    \label{fig:1a}
  \subfloat[Contextual Affordances ]{%
        \includegraphics[width=0.18\paperwidth,height=0.150\paperheight]{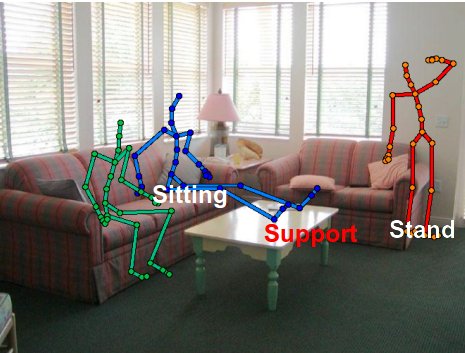}}
    \label{fig:1b}
      \subfloat[Affordance detection]{%
        \includegraphics[width=0.20\paperwidth,height=0.150\paperheight]{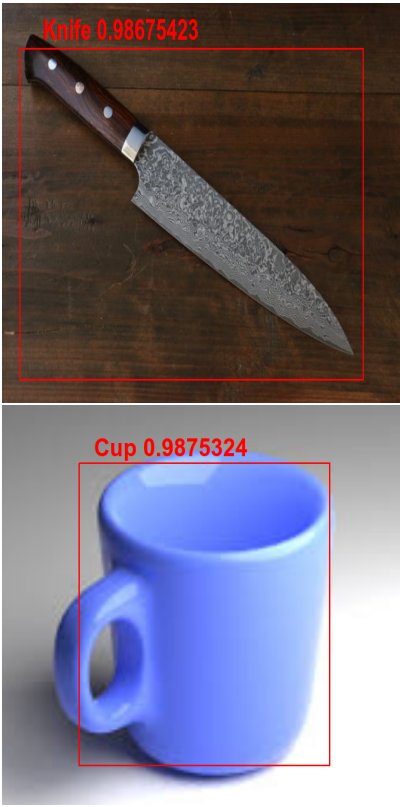}}
    \label{fig:1c}
  \subfloat[Affordance categorization]{%
        \includegraphics[width=0.23\paperwidth,height=0.150\paperheight]{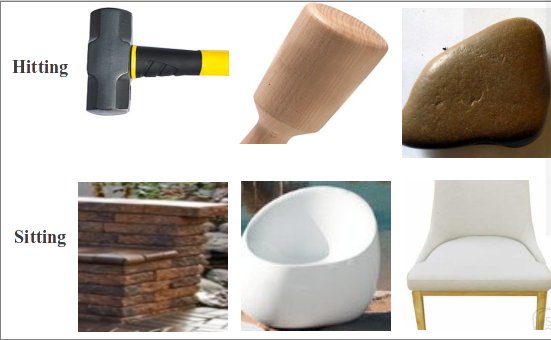}}
    \label{fig:1c}\
  \subfloat[Affordance segmentation ]{%
        \includegraphics[width=0.20\paperwidth,height=0.150\paperheight]{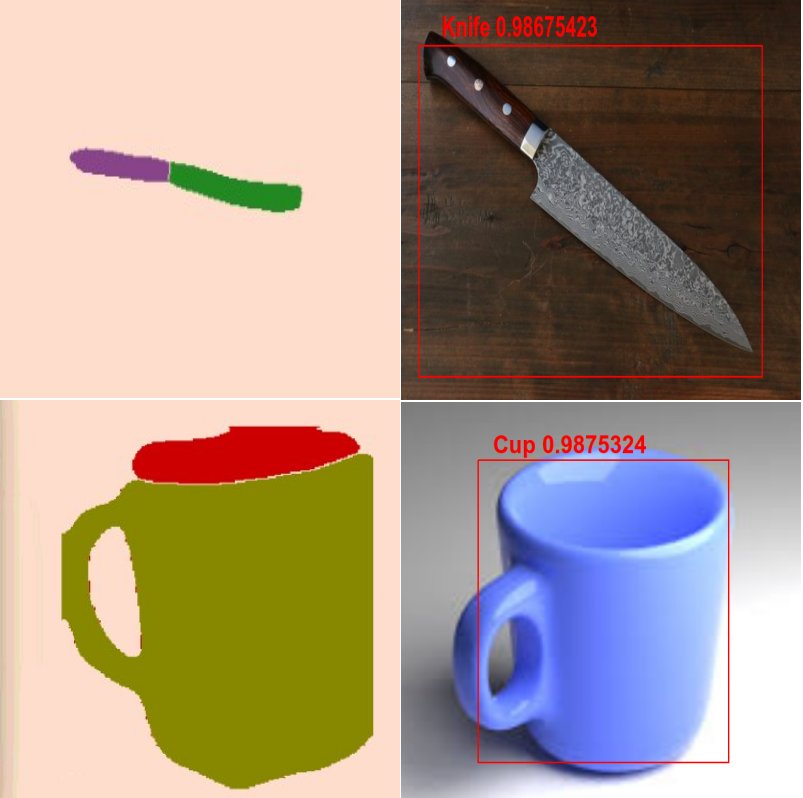}}
     \label{fig:1d} 
     \subfloat[Affordance reasoning]{%
	        \includegraphics[width=0.20\paperwidth,height=0.150\paperheight]{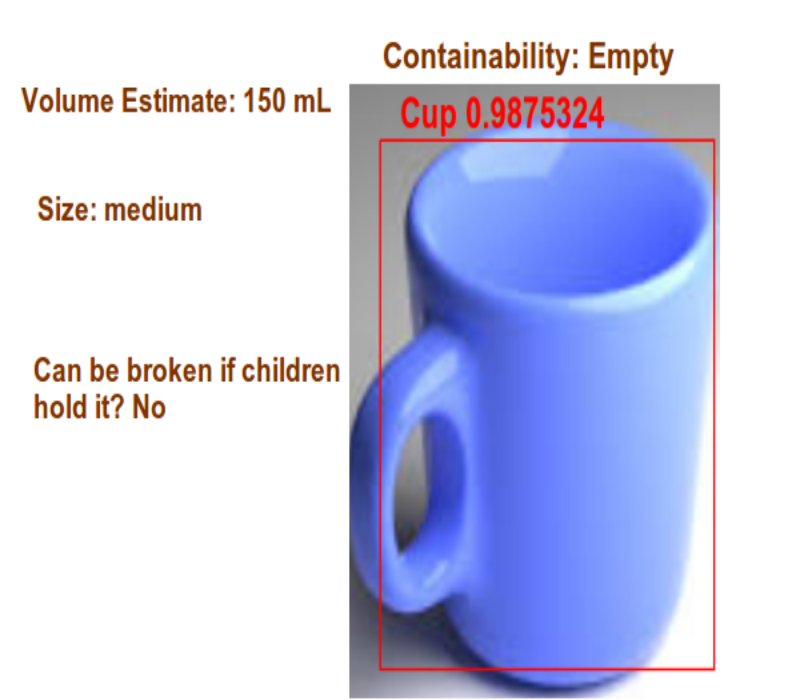}}
     \label{fig:1e} 
       \subfloat[Affordance-based activity recognition]{%
        \includegraphics[width=0.20\paperwidth,height=0.150\paperheight]{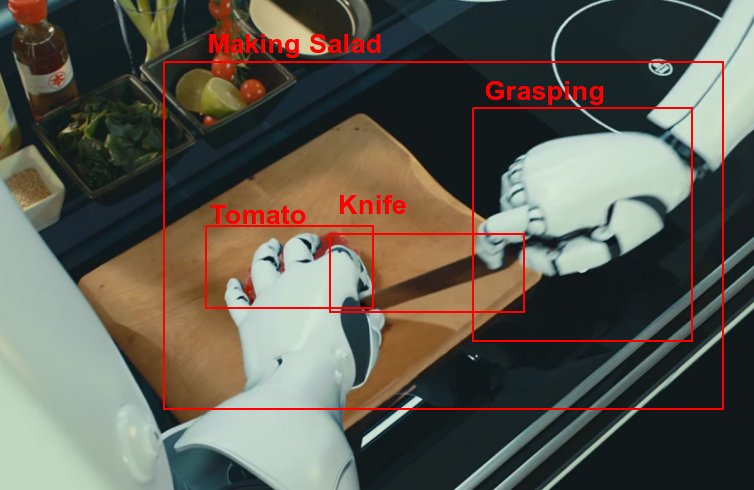}}
     \label{fig:1f} 
       \subfloat[Social affordances\cite{Learning_ActProperly_2017}]{%
        \includegraphics[width=0.23\paperwidth,height=0.150\paperheight]{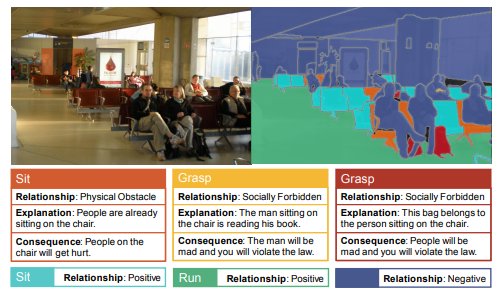}}
    \label{fig:1g}\
  \caption{(a),(b) Detecting functional parts to understand the scene usage and learn the affordances e.g. learning by observation. (c) Detecting the affordance objects and labels together using bounding boxes. (d),(e) affordance-based categorization and semantic labeling. (f) reasoning affordances from visual inputs such as how much water in the glass. (g), (h) given affordances, recognize the activities and social situations. }
  \label{fig1} 
\end{figure*}

\noindent
$-$\textbf{Functionality understanding:}
It transcends the traditional tasks of object detection and segmentation in visual scene understanding and aims to   understand the function of objects in a scene. For example, detecting the electricity plug that is required to charge the laptop or mobile phone as shown in Figure \ref{fig1} (a).

\noindent
$-$\textbf{Affordance learning:}
In the context of complex and dynamic environments, robots need to learn what can be done and what cannot? In other words, affordance learning involves teaching the robot to learn the possible set of actions that can be performed in a given environment. It addresses the possible effects that arise through object-agent actions. It circulates (overlaps) three main factors, that are object, action and effect as shown in Figure \ref{fig:objects_actions_Image}.

\noindent
$-$\textbf{Affordance detection:}
Similar to object detection, it localizes and labels the affordance for scene objects. Different from conventional detection tasks, it targets only the salient objects that are most relevant for actions (instead of all objects in the scene). The goal of detecting these affordances is to predict the next action or recognize the function of some objects, therefore it selectively targets only significant objects. Formally, let $X\,=\{x_{1},x_{2},...,x_{n}\}$ denotes the set of candidate bounding boxes with  $n$ instances and $Y\,=\{y_{1},y_{2},...,y_{n}\}$ denotes the output label space where $ y_i = \{(r_{i},l_{i}) : \: i \in [1,n] \}$ denotes a tuple consisting of instance location and affordance label respectively. Then, the affordance detection aims to find the function $f:X\rightarrow Y$.

\noindent
$-$\textbf{Affordance categorization:}
It means classifying the input images into a possible set of affordance categories. Often, this step is used as a precursor to affordance detection to make the process of localization and recognition more easier. Formally, let $I$ denotes an input image and $Y$ denotes its ground-truth affordance labels. The categorization task seeks to learn the optimal function $f:I\rightarrow Y$.

\noindent
$-$\textbf{Affordance reasoning:}
Affordance reasoning refers to more complex understanding of affordances which requires higher-order contextual modeling. The main purpose of such reasoning is to infer hidden variables like the amount of liquid inside a bottle or how much water can be pured into an empty bottle. Formally, let $X\,=\{x_{1},x_{2},...,x_{n}\}$ denotes the input space where $n$ is the number of object instances, $C\,=\{c_{1},c_{2},...,c_{n}\}$ denotes the contextual space (physical attributes, material properties, neighborhood and semantic relationships) and $Y\,=\{y_{1},y_{2},...,y_{n}\}$ denotes the output label space. The reasoning task is to find a function to map a relation such that $f:(C,X) \rightarrow Y$.

\noindent
$-$\textbf{Affordance semantic labeling:}
This task involves segmenting an image into a set of regions which are labeled with a semantically meaningful affordance category. Remarkably, this task assigns a category label to each pixel in a region of interest. Formally, let $P\,=\{p_{1},p_{2},...,p_{n}\}$ denotes the set of $n$ image pixels and  $Y\,=\{y_{1},y_{2},...,y_{n}\}$ denotes the output label space. The segmentation requires learning a function $f: P \rightarrow Y$ to assign a label $y$ to each image pixel.

\noindent
$-$\textbf{Affordance-based activity recognition:}
Objects bear possible actions of an agent which are represented as their affordances. Thus, recognizing the affordances is a crucial step towards complete activity recognition. This task aims to represent  the activities in terms of affordances. Note that we refer to atomic single-person operation as an `action', while actions performed by multiple people in a complex environment as an `activity', e.g., a moving robot is performing an action while a group of marching robots are performing an activity. Formally, let $X\,=\{x_{1},x_{2},...,x_{n}\}$ denotes the input space where $n$ is the number of object instances, $A\,=\{a_{1},a_{2},...,a_{n}\}$ denotes affordance space for each instance in the input space $X$ and  $Y\,=\{y_{1},y_{2},...,y_{n}\}$ denotes the
activities label space. Then, the activity recognition task is to learn optimal function $f:(X,A) \rightarrow Y$.

\noindent
$-$\textbf{Social affordances:}
Social affordance are a type of affordances that offer possible object-social actions. The term social imposes the human interaction through affordance learning. These social affordances may be positive such as sit on empty chair or negative (socially forbidden) to open a lady bag lying on the next chair. Formally, let $X\,=\{x_{1},x_{2},...,x_{n}\}$ denotes the input space where $n$ is the number of object instances, $E\,=\{e_{1},e_{2},...,e_{n}\}$ denotes objects interactions space for each instance in the input space $X$ and 
$Y\,=\{y_{1},y_{2},...,y_{n}\}$ denotes the activities label space. The activity recognition task is to find the function $f:(X,E) \rightarrow Y$.

\section{Significance}\label{Significance}
Affordance learning is a crucial task in computer vision and human machine interaction. Affordance relates to objects, actions and effects, therefore addressing it (i mean affordance) benefits all these associated fields. The significance of affordance and functionality understanding is summarized below:

\noindent
$-$\textbf{Anticipating and predicting future actions:}
Because affordances represent the possible actions that can be performed on or with objects, learning helps in action anticipation in a given environment. The examples in the literature that used affordances to predict future actions are \cite{Anticipating_human_activities_affordances_2016, Human_Activities_Affordance_2013, Structura_RNN_2016}.

\noindent
$-$\textbf{Recognizing agent's activities:}
The presence of an agent (e.g., a robot or a human) in the scene opens the possibility of its interactions with the surrounding environment. Recognizing its activities becomes an indispensable task to develop a complete understanding of scenes. Agent activities are highly dependent on the types of actions that are possible or more likely in a given environment. Several studies have targeted the problem of affordance-based human activity recognition \cite{Human_Activities_Grammar_2017, Earley_Parser_1970, Predicting_actions_static_scenes_2014}. 

\noindent
$-$\textbf{Provides valid functionality of the objects:}
Conventional object recognition task does not offer knowledge about its function and affordance, for example an occupied or a broken chair will still be recognized as a chair but a person cannot sit on it. In other words, recognizing what functions an object offers is a compulsory task to use it, particularly for the case of interactive robots \cite{WhatMakesChairAChair_2011}.

\noindent
$-$\textbf{Understanding social scene situations:}
In social situations, forbidden actions such as a healthy person occupying a disabled seat, happen daily in our life. However, it needs to be learned in case of visual recognition. Learning social affordances helps in understanding the whole scene \cite{Learning_ActProperly_2017}.

\noindent
$-$\textbf{Understanding the hidden values of the objects}
The recognition of an object's category does not provide the intuition about its value or significance. For instance, the task of knocking a nail would be normally done with a hammer. However, what if the hammer is not available? Learning affordances tells us that appropriate sized stones can also be used for the same task \cite{Understanding_tools_Affordances_2015}.

\noindent
$-$\textbf{Detailed scene understanding}
In traditional classification tasks, different kinds of pots are labeled with same label "pots". However, the usage of each pot could be different e.g., some could store rice while other could be suitable for vegetables. Therefore, categorizing objects according to their functionality is extremely important for detailed (or fine-grained) scene understanding \cite{Action_Recognition_Affordance_2011, WhatMakesChairAChair_2011}.

\noindent
$-$\textbf{Affordance cues benefit object recognition}
Extracting affordance cues from a scene provide the model with contextual information about objects that make the task of recognition easier. Many researchers used affordance values as contextual information to classify objects   \cite{Human_workspace_affordance_2011, RelativeAttributes_2011}

\section{Challenges}\label{Challenges}

It is worth noting that detection is the first step in the process of affordance learning. It is a crucial task and it should be accomplished carefully. It also involves several challenges, such as scale, illumination, appearance and viewpoint variations. Furthermore, because affordance detection is a kind of object detection, it inherits all the object detection problems as well. Some specific challenges are explained in detail below: 

\noindent
$-$\textbf{Illumination Conditions:}
Illumination changes affect the final results of affordance learning because they change the quality of the image being processed. Robot will be stuck if the electricity switches off suddenly particularly indoors building. Likewise, outdoor robots may face the same situation if the sun light changes suddenly from sunny to cloudy or even slowly from day to night. It affects the image at the pixel level, therefore any changes in the illumination will change the final accuracy. 

\noindent
$-$\textbf{Occlusion and Clutter:}
Occlusions often occur in dense images where one object obstructs parts of other objects. Background clutter adds difficulty to image recognition because of unordered objects. This sort of conditions leads the algorithm to incorrect predictions. Generally, this problem exists in multi-object images such as kitchens in indoor scenes and crowded scenes in the outdoor case. To allow the robot to perform its tasks easily, this problem needs to be addressed properly \cite{Occlusion_RelationalAffordance_2014,Occlusion_Manipulation_2013,Cluttered_Manipulations_2014}. 

\noindent
$-$\textbf{Viewpoint Variations:}
Images acquired from different viewpoints can affect the performance of recognition algorithms. Therefore, the orientation of an object with respect to the camera defines the accuracy of the method. For robots, robustness against viewpoint and pose changes is one of the main effective factors in the process of affordance learning \cite{Grasp_ViewPointVariations_2011}. 

\noindent
$-$\textbf{Scale Variations:}
Scale of the object in terms of the size is an important factor in the process of affordance learning especially with tools. For example, a fruit knife is generally small in comparison to a meat knife. Affordance detection algorithms should therefore be scale invariant to generalize well to unseen examples. 

\noindent
$-$\textbf{Deformation and intra-class variations:}
Deformation or the different shapes of the same object is another
criterion that needs to be treated to ensure reliable actions \cite{Shape_Variation_2008}. 

\noindent
$-$\textbf{Single Object Multiple Labels (SOML):}
The characteristics of affordance learning problems is different from traditional problems which have a single label for every instance or scene \cite{MultiLabel_tutorial_2015,MultiLabel_Review_2014}. For example, a knife in the kitchen has a grasping label in the hand and a cutting label in the cutter. Similarly, a cup of tea has a grasping label outside and a pouring label inside. Hence, visual affordance inherit all the problems of multi-label learning paradigm such as ranking, correlation, dependency and multi-label scheme representation.

\noindent
$-$\textbf{Multiple Objects Multiple Labels (MOML):}
In contrast to multi label problems, affordance cases have multiple labels at the object level rather than the complete scene level. It is closer to multi-instance multi-label (MIML) problems if we consider the objects as instances \cite{Multi-instance_multi-label_learning_2012,Multi-instance_multi-label_learning_scene_Classification_2012,Multi-instance_Learning_2017}. Intuitively, MOML inherits MIML challenges as well as single object multiple label difficulties, it has to address higher-order correlations between instance, dependency as well as the exponential size bottleneck.

\begin{figure}[htp]
\centering \includegraphics[width=0.8\columnwidth]{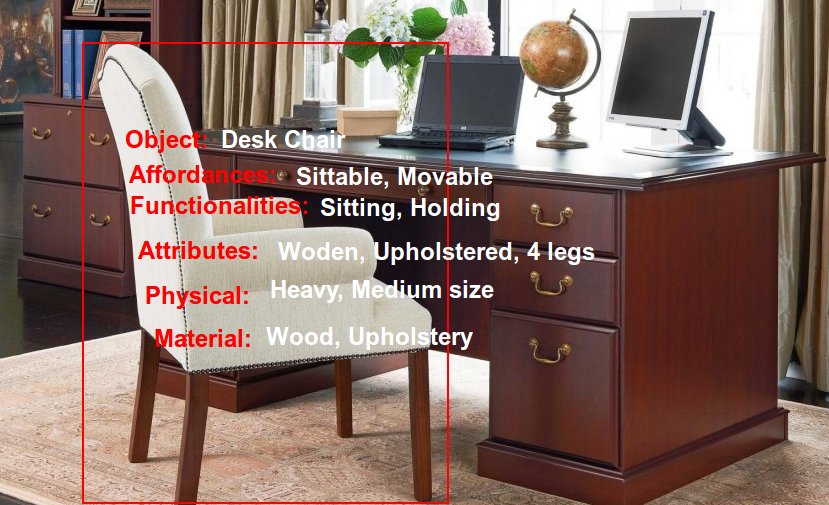}
\caption{Multiple sources of features that can be fused together to create the affordance context. }\label{fig:multisource}
\end{figure}

\noindent
$-$\textbf{Multi-source features:}
The features required to perform affordance learning are diverse in nature and should come from complementary sources \cite{Affordance_PhysicalProperties_2007} (see Figure \ref{fig:multisource}): 
\begin{enumerate}
\item \textbf{Visual cues} such as color and texture.
\item \textbf{Physical  properties} such as weight and volume,
for instance, the chair's visual feature is movable, but the weight, which is a physical property, does not allow sitting. 
\item \textbf{Material properties} e.g., to assess the comfort of a chair. 
\end{enumerate}

\noindent
$-$\textbf{Inter-object dependency:}
The affordance of an object sometimes relies on other objects, particularly in service robots. For example, for the task of preparing a cup of tea, the robot has to boil water which depends on the electricity plug and pour hot water in a cup. This task will be complicated if one of those objects is  unseen or occluded. \cite{Affordance_Challenges_2016}. 

\noindent
$-$\textbf{Human-Objects dependency:}
The existence of a human in a scene makes it more complex
in terms of actions, events and affordance learning in particular because the affordance is then dependent on the attributes of the human. For example, putting some fruit in the fridge depends on the height of the human.

\section{What is a Visual Affordance?}\label{Visual Affordance}
The field of scene understanding aims to allow computers to be able to understand the environment and its contents. It has been studied from different perspectives: object recognition \cite{RCN_2015,Imagenet_2015}, object detection \cite{YOLO_redmon2016, SSDliu2016,Detection_Poselets_2009, Detection_DPM_2010, Detection_Selective_Search_2013},  scene classification \cite{Scene_Classification_2015,Scene_Classification_2014}, indoor scene understanding \cite{Indoor_scene_understanding_2015,Indoor_Scene_Understanding_2015_1} and so on. Much research has been conducted to address these problems. However, understanding possible interactions and developing a higher level reasoning about scenes has been less investigated. In other words, merely detecting the scene elements is not enough to make intelligent decisions but inferring complex interactions and dynamics in a scene: What the possible actions that can me made? For example, learn how to use the vacuum cleaner inside the kitchen? Which button should be pressed? Where is the electricity plug?  What is required to name the chair a chair? It may not be sittable, occupied or broken \cite{WhatMakesChairAChair_2011}. To sum up, maximizing the benefits of scene understanding requires other associated factors such as affordance detection and reasoning to make best use of it.

\begin{figure}
\centering \includegraphics[width=0.8\columnwidth]{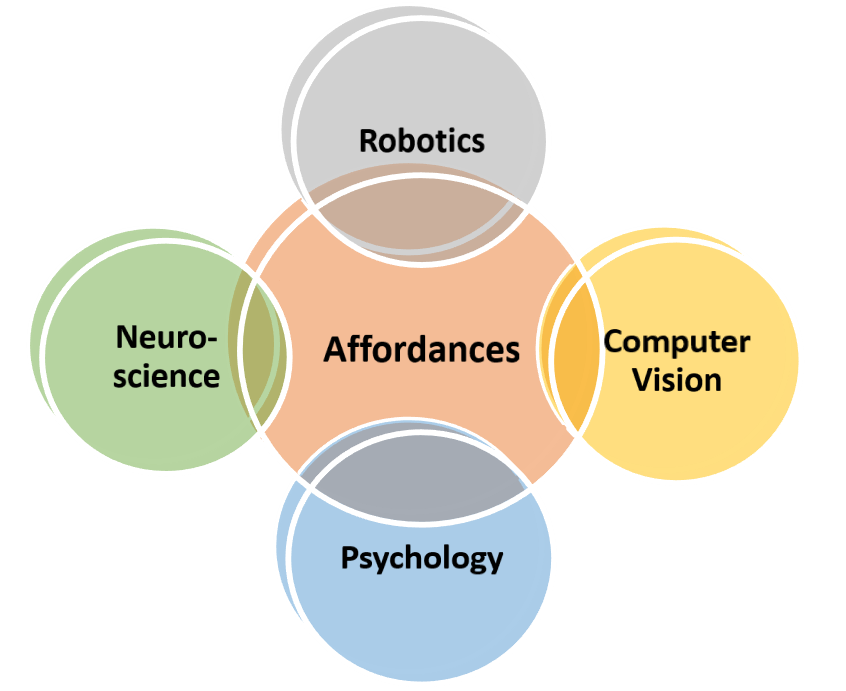}\label{fig:AffordanceRelations}
\caption{Affordances and other fields}
\end{figure}

The concept of affordance has initially been engaged strongly in the fields of perceptual psychology, cognitive psychology and environmental psychology. Affordances are often used for testing the capability of objects' interactions because it has tight relations with different environment types, it addresses various research areas and it is linked to the object where ever it is. In detail, affordances differ according to the environment. For instance, the robot inside the kitchen should understand the affordance of tools \cite{AffDetectGeometricFeatures_2015}, CAD objects such as chairs \cite{Aff_Function-based_1994} and the functionalities that can be done such as the functionality of electricity plug to run the vacuum cleaner \cite{DFSU2017}. Moreover, if the human is involved in the scene, a new aspect, human-robot interaction and action recognition, should be addressed. After that it has been introduced to the fields of computer vision, human-robot interaction and robotics \cite{AfforFirstUsedVision_2007}. The Affordance field is inter-related to different fields such developmental learning, robot manipulation and psychology. In this survey, the visual affordance will be summarized because of the lack of any existing literature review in this area.

\begin{figure}
\centering \includegraphics[width=\columnwidth]{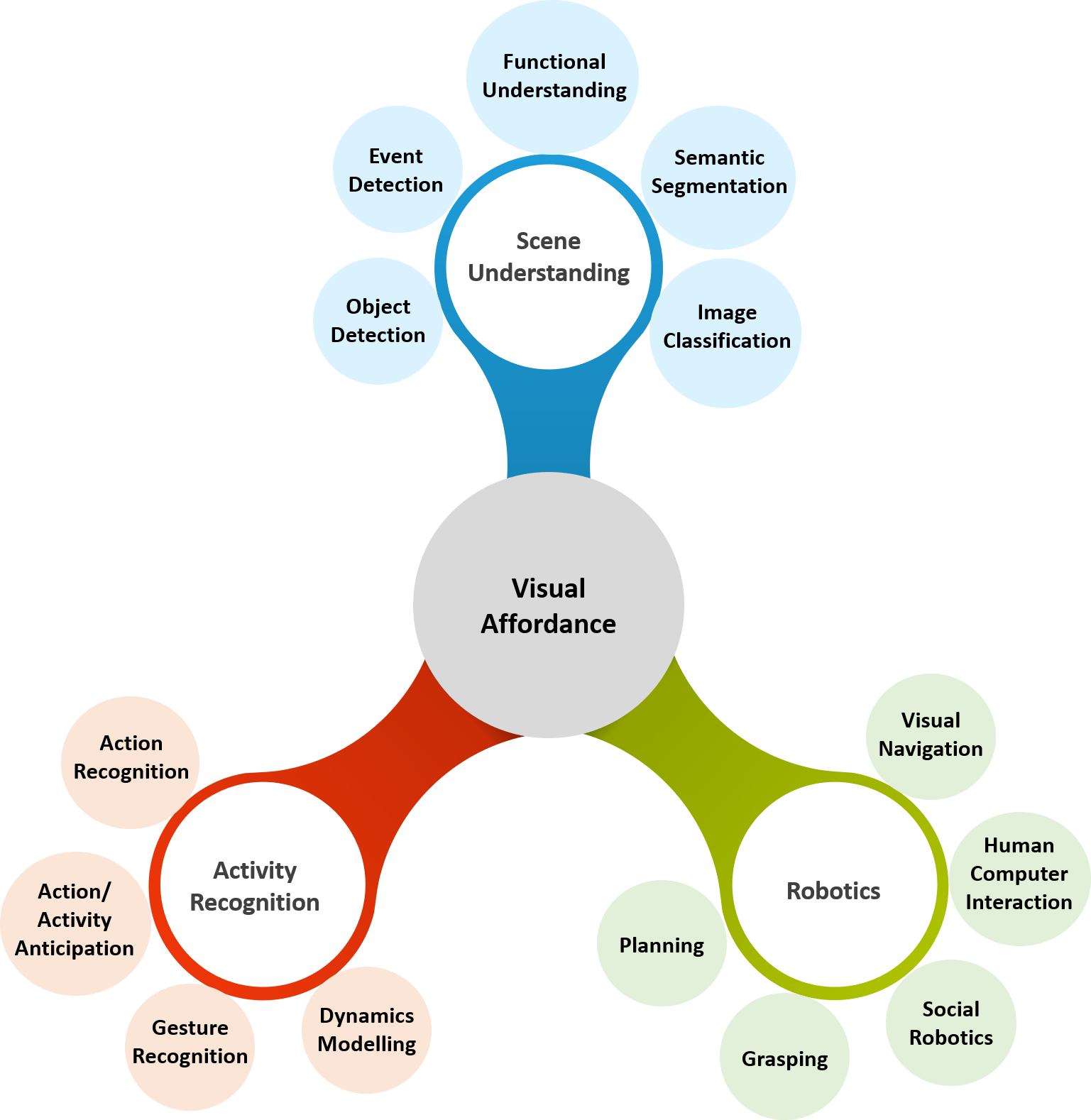}\label{fig:VisualAffordances}
\caption{Relation between affordances and visual fields}
\end{figure}

Visual affordance is a branch of research that deals with the affordances as computer vision problem based on images and videos; and uses the machine learning methods, such as deep learning, to solve these challenges. It is tightly connected to various fields: action recognition \cite{Action_Recognition_Affordance_2011,Human_Activities_Affordance_2013}, scene understanding, grasping, human-robot interaction \cite{Actions_Affordance_Mining_Semantic_2015,HRI_SocialAffordances_2016} and function recognition \cite{DiscoverObjectFunction_2013,DFSU2017}. Therefore it overlaps with important computer vision problems such as image classification and action recognition; and affected by identical set of challenges such as illumination changes, occluded scenes, pose of objects and dynamic scenes \cite{Scene_Understanding_Survey_2017}. Apart from that, the problem of affordance learning is more complicated because on top of all inherited difficulties from computer vision, it encounters additional challenges e.g. each object may have different labels/functionalities.

\subsection{Affordance Detection}\label{subsection:challenges}

\begin{figure*}
\centering \includegraphics[width=\textwidth]{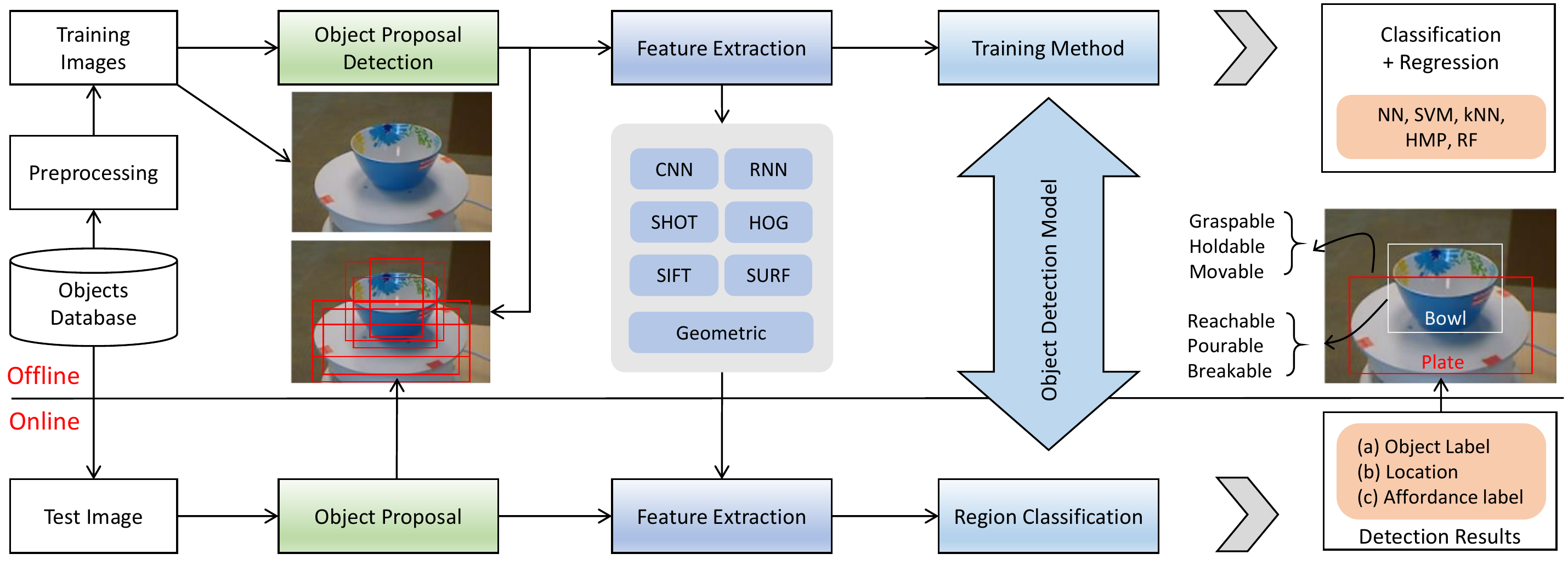}
\caption{Detection process through machine learning techniques.}
\label{fig:detectionFlowchart}
\end{figure*}

Recent advances in robotics and computer vision have paved the way for autonomous agents to impact every aspect of our life. To allow service robots do tasks, they need to understand the environment. For example, the robot cannot use the hammer without localizing the handle. Therefore, detecting the affordance of the scene, tool or a given image is a mandatory task for effective interaction and manipulation (see Figure \ref{fig:detectionFlowchart}). This task is, however, quite challenging, as explained below.

\subsubsection{\textbf{Feature-engineering approaches}}
The affordance detection task deals with labeling and localization of particular parts that can afford certain actions. It also involves reasoning about the functionality of a tool. Early work in affordance detection sought to recognize 3D CAD objects e.g., chairs based on the object functions. Stark and Bowyer \cite{Aff_Function-based_1994} built the Generic Recognition Using Form and Function (GRUFF) system to recognize different objects according to the functionalities rather than the shapes. They used several functional primitives which were assigned to perform function-based recognition. Aldoma \etal \cite{zero_order_affordances_detection_2012} proposed a visual cue method to find affordances in the scene depending on the pose of the object. Their method depends on first recognizing the objects in a scene and then estimating its 3D pose. Finally, they learn the so-called 0-Order affordances which refers to hidden and unhidden affordances (see Table \ref{table:affordances} for the investigated affordances). 

\begin{table*}
\begin{tabular*}{1\textwidth}{@{\extracolsep{\fill}}>{\raggedright}m{.15\textwidth}>{\raggedright}m{0.08\textwidth}>{\raggedright}m{0.35\textwidth} >{\raggedright}m{0.20\textwidth}>{\raggedright}m{0.2\textwidth}}

\hline 
Affordance Label & References & Description & Examples &   Method \tabularnewline
\hline 

rollable  & \cite{Visual_Physical_Affordances_2011} ,  \cite{learning_to_Label_Affordances_2017} ,  \cite{zero_order_affordances_detection_2012}  & indicate whether the object is rollable or not & roads, trolley & detection\tabularnewline
\vspace{2mm}
containment  & \cite{Weakly_Supervised_Learning_Affordance_2016} ,\cite{AffDetectGeometricFeatures_2015},  \cite{zero_order_affordances_detection_2012}, \cite{Object_based_Affordances_CNN_CRF_2017, Bootstrapping_semantics_tools_2016} & indicates contain-ability of an object & pots & detection \tabularnewline
\addlinespace[2mm]
liquid-containment &   \cite{zero_order_affordances_detection_2012} & indicates liquid contain-ability of an object & glasses, cups, mug& detection \tabularnewline
\addlinespace[2mm]
unstable  &   \cite{zero_order_affordances_detection_2012} & indicates whether the object pose is stable after pushing or not & glass cups will be broken in case of pushing& detection
\tabularnewline
\addlinespace[2mm]
stackable-onto &   \cite{zero_order_affordances_detection_2012} & indicates that the object can be stacked & mugs, pots& detection\tabularnewline
\addlinespace[2mm]
sittable &   \cite{zero_order_affordances_detection_2012}, \cite{learning_to_Label_Affordances_2017}, \cite{multi_scale_affordance_segmentation_2016}, \cite{ Semantic_Labeling_3D_Clouds_2014}& indicates whether the object can be used to sit or not & chairs, desks& detection, segmentation\tabularnewline
\addlinespace[1mm]
grasp & \cite{AffDetectGeometricFeatures_2015},  \cite{Visual_Physical_Affordances_2011} , \cite{Object_based_Affordances_CNN_CRF_2017}, \cite{learning_to_Label_Affordances_2017} , \cite{ Semantic_Labeling_3D_Clouds_2014, Affordance_Sensorimotor_Recognition_2017} & indicates the location of manipulation of flat tools & hammer, cups & detection\tabularnewline

cut  & \cite{AffDetectGeometricFeatures_2015}, \cite{Weakly_Supervised_Learning_Affordance_2016} , \cite{Object_based_Affordances_CNN_CRF_2017, Affordance_Sensorimotor_Recognition_2017,Bootstrapping_semantics_tools_2016} & indicates cutting  & knife, penknife, key& detection\tabularnewline

scoop  & \cite{AffDetectGeometricFeatures_2015} & indicates curved surfaces tools & trowels, cookie scoop,  gutter scoop& detection\tabularnewline
\addlinespace[2mm]
pound  & \cite{AffDetectGeometricFeatures_2015} , \cite{Object_based_Affordances_CNN_CRF_2017}& indicates striking tools & hammer head& detection\tabularnewline
\addlinespace[2mm]
support,place-on & \cite{AffDetectGeometricFeatures_2015} , \cite{Weakly_Supervised_Learning_Affordance_2016} , \cite{Object_based_Affordances_CNN_CRF_2017} & indicates flat tools, helpers or support an agent & flat tools (turners,spatulas), place-on (tables, desks), agent support (walls)
as walls & detection\tabularnewline
\addlinespace[2mm]
wrap-grasp & \cite{AffDetectGeometricFeatures_2015} , \cite{Object_based_Affordances_CNN_CRF_2017} & indicates the location of grasping of rounded tools like cups& the outside of a cup) & detection\tabularnewline
\addlinespace[2mm]
pushable, pushable forward, pushable right, pushable left  & \cite{Visual_Physical_Affordances_2011}, \cite{ Semantic_Labeling_3D_Clouds_2014, Affordance_Sensorimotor_Recognition_2017}  & indicates whether the object is push-able of an object & trolley, bike& detection,segentation\tabularnewline
\addlinespace[2mm]
liftable & \cite{Visual_Physical_Affordances_2011}, \cite{ Semantic_Labeling_3D_Clouds_2014, Affordance_Sensorimotor_Recognition_2017}  & indicates whether the object can be lifted or no & liftable chairs  & detection, segmentation\tabularnewline
\addlinespace[2mm]
dragable, pushable backward & \cite{Visual_Physical_Affordances_2011}, \cite{ Semantic_Labeling_3D_Clouds_2014} & indicates whether the object can be dragged & desk, table& detection, segmentation\tabularnewline
\addlinespace[2mm]
carryable  & \cite{Visual_Physical_Affordances_2011}  & indicates whether the object can be carried& light-weight pots, balls& detection\tabularnewline
\addlinespace[2mm]
traversable & \cite{Visual_Physical_Affordances_2011} & indicates whether the object can be traversed & road, grass& detection\tabularnewline
\addlinespace[2mm]
openable  & \cite{Weakly_Supervised_Learning_Affordance_2016, Affordance_Sensorimotor_Recognition_2017} & indicate whether the object can be opened& fridge, room, microwave, book, box & detection\tabularnewline
\addlinespace[2mm]
pourable & \cite{Weakly_Supervised_Learning_Affordance_2016}  & indicates whether the object is pour-able & mug & detection\tabularnewline
\addlinespace[2mm]
holdable &  \cite{Weakly_Supervised_Learning_Affordance_2016}& indicates whether the object can be hold& the outside of mug & detection\tabularnewline
\addlinespace[2mm]
display, observe &  \cite{Object_based_Affordances_CNN_CRF_2017},  \cite{learning_to_Label_Affordances_2017} & refers to display sources & TV , monitor screen & detection\tabularnewline
\addlinespace[2mm]
engine & \cite{Object_based_Affordances_CNN_CRF_2017} & refers to tool\textquoteright s engine parts & drill engine & detection \tabularnewline
\addlinespace[2mm]
hit  & \cite{Object_based_Affordances_CNN_CRF_2017, Bootstrapping_semantics_tools_2016} & refers to tools could be used to strike other objects.
&racket head& detection\tabularnewline
\addlinespace[2mm]
obstruct & \cite{learning_to_Label_Affordances_2017}& indicates the locations of obstructer &wall & detection\tabularnewline
\addlinespace[2mm]
break  & \cite{learning_to_Label_Affordances_2017}& indicates break-sensitive objects &  glass cups& detection\tabularnewline
\addlinespace[2mm]
pinch-pull   & \cite{learning_to_Label_Affordances_2017} & indicates objects that should be pulled with punch &  knob& detection\tabularnewline
\addlinespace[2mm]
hook-pull  &  \cite{learning_to_Label_Affordances_2017} & indicates objects that should be pulled with hooking up & handle& detection\tabularnewline
\addlinespace[2mm]
tip-push  & \cite{learning_to_Label_Affordances_2017} & indicates objects that perform actions after pushing & electricity
buttons& detection\tabularnewline
\addlinespace[2mm]
warmth  & \cite{learning_to_Label_Affordances_2017}& indicates warmth objects & fireplaces& detection\tabularnewline
\addlinespace[2mm]
illumination & \cite{learning_to_Label_Affordances_2017} & indicates light objects & lamps & detection\tabularnewline
\addlinespace[2mm]
dry   & \cite{learning_to_Label_Affordances_2017}& indicates objects that absorb water  &towels& detection\tabularnewline
\addlinespace[2mm]
walk & \cite{learning_to_Label_Affordances_2017} , \cite{multi_scale_affordance_segmentation_2016}& indicates places that allow walking & gardens  & detection, segmentation\tabularnewline
\addlinespace[2mm]
lyable  & \cite{multi_scale_affordance_segmentation_2016} &refers to long free space that allow person to lie down & bed & segmentation\tabularnewline
\addlinespace[2mm]
reachable &\cite{multi_scale_affordance_segmentation_2016}  & refers to object in a scene that is reachable for a person to pick it& water bottle from the fridge & segmentation
 \tabularnewline
\addlinespace[2mm]
movable &\cite{Semantic_Labeling_3D_Clouds_2014} & refers to objects that can be moved around& small objects like balls, mugs&segmentation\tabularnewline
\addlinespace[2mm]
\hline 
\end{tabular*}
\caption{Indoor affordance labels used to detect objects  as in studies \cite{zero_order_affordances_detection_2012, AffDetectGeometricFeatures_2015, Visual_Physical_Affordances_2011, Weakly_Supervised_Learning_Affordance_2016, Object_based_Affordances_CNN_CRF_2017, learning_to_Label_Affordances_2017, multi_scale_affordance_segmentation_2016,  Semantic_Labeling_3D_Clouds_2014, Affordance_Sensorimotor_Recognition_2017,Bootstrapping_semantics_tools_2016}}
\label{table:affordances} 
\end{table*}

Myers \etal \cite{AffDetectGeometricFeatures_2015} were the first  to treat images from the perspective of geometry as traditional vision tasks. They detected affordance of tool parts based on its importance for robot vision. They introduced RGB-D data set with ground truth annotations whereas SVM was the learning algorithm. This study used a pixel-wise method to generate geometrical features to learn through it, but they used hand-crafted features. The idea behind using pixel-wise methods was brand new in this
sense, however, it is complicated because the same pixel may share
different affordance labels. They used two methods to to train this model. Firstly, S-HMP (Superpixel-based Hierarchical Matching Pursuit) \cite{HMP_2013} to extract geometric features (depth,
surface normals, principle curvatures , and shape-index and curvedness) and SVM as the main classifier. In addition, they used S-RF (Structured Random Forest) \cite{Structured_forests_2013} to infer the affordance labels based on extracted decision trees particularly in real-time basis. They introduced seven affordance labels
as shown in Table \ref{table:affordances}.  

Herman \etal \cite{Visual_Physical_Affordances_2011} sought to introduce a new method which depends on physical and visual features such as material, shape, size and weight to learn the affordances labels (as shown in Table \ref{table:affordances} ). They collected their own data, which belongs to six categories: balls, books, boxes containers (mugs, bottles, and pitchers) shoes and towels, using a mobile robot Pioneer 3 DX. Based on these features, they used SVM and k-nn classifiers to test their method. Their work emphasized the concept that combining physical and visual attributes together  enhances the affordance learning. 

Grabner et al. \cite{WhatMakesChairAChair_2011} used the concept
of functionality to recognize whether an object (e.g. a chair)
is sittable or not. In other words, the chair allows the sitting affordance, but it may be used by another object. In order to detect whether the chair affords sitting or is \textquotedbl{}sittable\textquotedbl{}, they used a human skeleton 3D model to test sitting in 3D models of chairs. Through different sitting human poses; and interaction between an actor and object, they detect not only a chair's affordances but also how to use it. Despite its effect in detection and reasoning, it required additional cues to improve the performance.

Moldovan \etal \cite{Occlusion_RelationalAffordance_2014} proposed a novel method to estimate objects affordances in an occluded environment. They used the relational affordances concept to search for objects which can afford certain actions \cite{Learning_relational_affordance_2012}. Additionally, they used Statistical Relational Learning (SRL) \cite{SRL_relational_learning_2008,SRL_relational_learning_2007} methods to model probability distributions that encode the relationships between objects. However, they generate these probabilities through an external knowledge base i.e., web images. Hassan and Dharmaratne \cite{Affordance_Challenges_2016} proposed an affordance detection method based on the object, human and the ambient environment. They used the objects attributes (physical, material, shape, etc), human
attributes (poses) and object-to-object to train their scheme. Local feature have been extracted and used as inputs for classification based on Bayesian networks.

\subsubsection{\textbf{Feature-learning approaches:}}
Inspired by \cite{AffDetectGeometricFeatures_2015}, Nguyen et al. \cite{Affordance_Detection_CNN_2016} built their model to extract geometric deep features using Convolutional Neural Network (CNN) to detect affordances. They used an encoder-decoder architecture based on deep CNNs with multi-modal features (horizontal disparity, height and angle between pixel normals and inferred gravity) \cite{MultiModal_2011,RichFeatures_2014}. It was demonstrated that the automatic feature learning performed on top of geometric features resulted in better performance compared to  \cite{AffDetectGeometricFeatures_2015}. In contrast, semantic segmentation \cite{Semantic_Segmentation_2015} is introduced to treat the affordance pixels. They tested it by real robot for grasping and they conducted their study on the UMD dataset \cite{AffDetectGeometricFeatures_2015}. Despite their significant enhancements, the data set images are simple i.e. it has no occlusion nor clutter. 
Sawatzky \etal \cite{Weakly_Supervised_Learning_Affordance_2016, Binarization_Affordance_segmentation_2017} proposed weakly supervised method to learn affordance detection using deep CNN based expectation maximization framework. This framework adequately handles weakly-labeled data at the image-level or key-point level annotations. They sought to fix the problem of affordance segmentation which
needs special care because every pixel may be assigned to
multiple labels. They learned deep features from the training data,
however, they used human pose to represent the context. They introduced
affordance RGBD datasets with rich contextual information. The
annotated labels are shown in Table \ref{table:affordances}.
However, the authors employed an additional step to update the parameters
of CNN and estimate required masks for segmentation \cite{Binarization_Affordance_segmentation_2017}.
For this reason, they proposed the adaptive binarization threshold approach
to get rid of that step and enhance the results. 
Nguyen \etal \cite{Object_based_Affordances_CNN_CRF_2017}  conducted
another study to detect the affordance labels in the images using
a deep CNN architecture. Similar to Ye \etal \cite{DFSU2017}, their work was inspired by popular CNN based object detectors \cite{DiscoverObjectFunction_2013,YOLO_redmon2016,Detection_Selective_Search_2013}, Resultantly they treated affordance learning as an object detection problem. This approach starts with a candidate set of  bounding box proposals for objects which is generated using \cite{R-FCN_2016,FasterR-CNN_2015}. Although They tested through Faster R-CNN \cite{FasterR-CNN_2015} and R-FCN \cite{R-FCN_2016} with various popular network architectures like VGG-16 \cite{VGG16_simonyan2014}, ResNet-51 and ResNet-101 \cite{Resnet_2016}, R-FCN outperformed the Faster R-CNN by a slight margin. This detection stage is followed by atrous convolution technique \cite{deeplab_2015} to extract deep features which are finally followed by a Conditional Random Fields (CRF) model. The CRF, as a post-processing mechanism, provides substantial improvements \cite{deeplab_2015,Deep_CRF_2015}. The authors published a new RGB-D dataset called IIT-AFF (with nine affordance labels as given in Table \ref{table:affordances}) to prove the efficacy of their method. The collected images in the dataset have good quality i.e., there do not exist many occlusions, cluttered regions and low resolution images. The authors in \cite{Object_based_Affordances_CNN_CRF_2017} used most recent architectures in deep learning, however, it resulted in high computational complexity due to high number of parameters in these models \cite{Detectors_2016}.
\begin{figure}
\centering \includegraphics[width=0.85\columnwidth,height=0.11\paperheight]{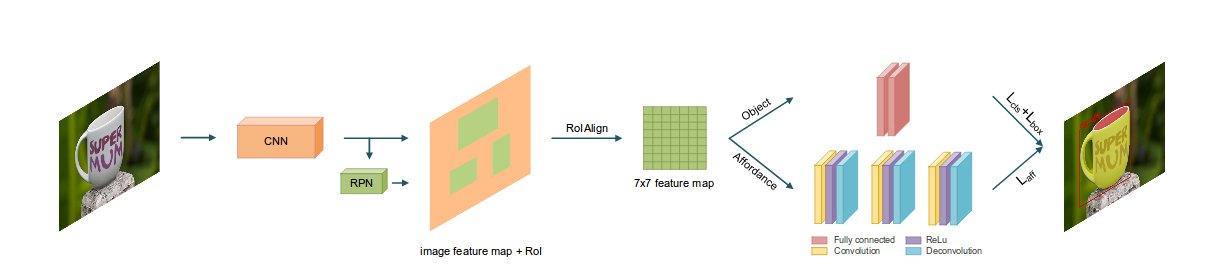}\caption{\label{fig:affordanceNeta} The deep learning architecture that is proposed by \cite{affordancenet2017_Do}. The concept of RoIs have been used \cite{Mask-R-CNN_2017_ICCV}  to share the weights with the main CNN layers whereas the VGG is the feature extractor. In addition, they used deconvolution layer to refine the results of affordances.}.
\end{figure}
Do \etal \cite{affordancenet2017_Do} followed the two previous studies of the same group \cite{Object_based_Affordances_CNN_CRF_2017,Affordance_Detection_CNN_2016} with a new effort to build an end-to-end deep learning architecture. Notably, the end-to-end model learning concept has recently predominated the recognition techniques, which has led to algorithms that can perform model training in a single framework \cite{SSDliu2016,YOLO_redmon2016}. To elaborate further, these systems detect the objects and their affordances in a single stage instead of multiple isolated and disintegrated steps. Hence, they reduce the training time and lead to better performances.  Although they mainly followed the same strategy of \cite{Mask-R-CNN_2017_ICCV,end-to-end_instance_segmentation_2017} as shown in Figure \ref{fig:affordanceNeta}, they added new components like deconvolutional layers and robust resizing strategies to handle the multiple affordance classes problem. They relied on the affordance labels of \cite{Object_based_Affordances_CNN_CRF_2017} and tested their approach on two datasets: UMD dataset \cite{AffDetectGeometricFeatures_2015} and IIT-AFF \cite{Object_based_Affordances_CNN_CRF_2017}. 
In contrast to the above mentioned methods, Chen \etal \cite{DeepDriving_Affordances_2015} proposed a new out of the box idea to utilize affordance learning to reason about autonomous driving actions. They trained deep Convolutional Neural Network (CNN) on the KITTI dataset \cite{KITTI_Dataset_2013} and twelve recorded hours
of video game \cite{Torcs} (see Figure \ref{fig:deepdriving}). The authors introduced 13 affordance indicators as
shown in Table \ref{table:DeepDriving} while, learning these indicators depend on the lane of that car and its perception. These learned affordances are tested to predict the right action. Eventually, a detailed comparison between affordance detection techniques is presented in Table \ref{table:affordancedetectioncomparison}.

\begin{figure}
\centering \includegraphics[width=0.85\columnwidth,height=0.11\paperheight]{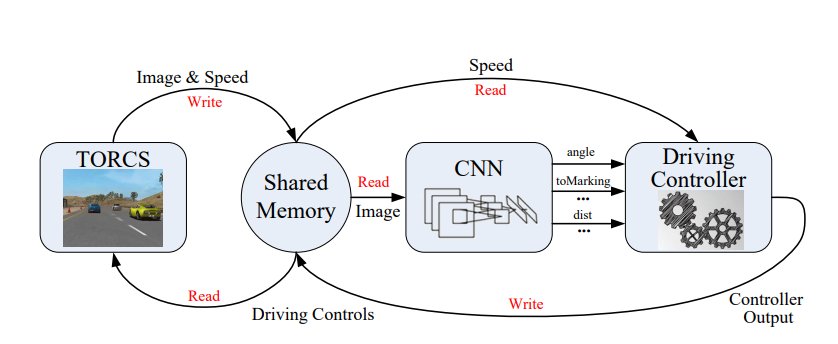}\caption{\label{fig:deepdriving} \textbf{The deep driving architecture} \cite{DeepDriving_Affordances_2015}. Given an image from TORCS \cite{Torcs}, the CNN extracts 13 indicators that will be fused along with the car speed to reason the proper command for this case}.
\end{figure}	
\begin{table}
\centering
\begin{tabular}{ll}
\hline 
\textbf{Affordance Indicator}  & \textbf{Description or Function} \tabularnewline
\hline 

angle  &angle between the car and road's tangent \tabularnewline

\multicolumn{2}{l}{ \textbf{When driving in the lane called \textquotedbl{}in lane system\textquotedbl{}}}\tabularnewline

to\_Marking\_LL  & marking distance between left \& left lane   \tabularnewline

to\_Marking\_ML  & marking distance between it \& left lane \tabularnewline

to\_Marking\_MR  & marking distance between  lane \& right lane \tabularnewline

to\_Marking\_RR  & marking distance between right  \& right lane \tabularnewline

dist\_LL  & distance between left lane \& preceding car \tabularnewline

dist\_MM  &distance between its lane \& preceding car\tabularnewline

dist\_RR  &distance from right lane to the preceding car \tabularnewline

\multicolumn{2}{l}{\textbf{When driving in the lane called \textquotedbl{}in marking
system\textquotedbl{}}}\tabularnewline

to\_Marking\_L &distance to the left lane marking \tabularnewline

to\_Marking\_M & distance to the central lane marking \tabularnewline

to\_Marking\_R & distance to the right lane marking \tabularnewline

dist\_L  &distance to the preceding if it is in the left \tabularnewline

dist\_R  & distance to the preceding if it is in the right\tabularnewline
\hline 
\end{tabular}
\caption{The outdoor affordance labels that are proposed by Chen et al. \cite{DeepDriving_Affordances_2015}}
\label{table:DeepDriving} 
\end{table}

\begin{table*}

\begin{tabular*}{1\textwidth}{@{\extracolsep{\fill}}>{\raggedright}m{0.1\textwidth}>
{\raggedright}m{0.02\textwidth}>
{\raggedright}m{0.02\textwidth} >
{\raggedright}m{0.02\textwidth}>
{\raggedright}m{0.09\textwidth}>
{\raggedright}m{0.09\textwidth}>
{\raggedright}m{0.02\textwidth}>
{\raggedright}m{0.02\textwidth}>
{\raggedright}m{0.02\textwidth}>
{\raggedright}m{0.02\textwidth}>
{\raggedright}m{0.02\textwidth}>
{\raggedright}m{0.02\textwidth}>
{\raggedright}m{0.02\textwidth}>
{\raggedright}m{0.02\textwidth}}
\hline 
\addlinespace[2mm]
 \multicolumn{1}{l|}{}& \multicolumn{3}{c|}{Object Features} & \multicolumn{2}{c|} {Features Extraction} & \multicolumn{3}{c|}{Evaluation}  & \multicolumn{3}{c|}{Training} & \multicolumn{2}{c}{Model}\tabularnewline
 \addlinespace[2mm]
\hline 

 \addlinespace[2mm]
 \multicolumn{1}{l|}{}&  \rotatebox[origin=c]{90}{2D}  & \rotatebox[origin=c]{90}{3D} &  \multicolumn{1}{c|}{\rotatebox[origin=c]{90}{Multimodal}} & \rotatebox[origin=c]{90}{Feature Learning}  &  \multicolumn{1}{c|}{\rotatebox[origin=c]{90}{Hand-crafted}} &  \rotatebox[origin=c]{90}{Real Robot}  & \rotatebox[origin=c]{90}{Simulation} & \multicolumn{1}{c|}{\rotatebox[origin=c]{90}{Benchmark}}  &  \rotatebox[origin=c]{90}{Supervised}  & \rotatebox[origin=c]{90}{Unsupervised}  & \multicolumn{1}{c|}{\rotatebox[origin=c]{90}{Weakly Supervised}} &  \rotatebox[origin=c]{90}{\parbox{0.7in}{Statistical/ Mathematical}} & \rotatebox[origin=c]{90}{Neural Net}\tabularnewline
  \addlinespace[2mm]

\hline 

\multicolumn{1}{l|} {Stark et al. \cite{Aff_Function-based_1994}}  &  & \checkmark  & \multicolumn{1}{c|} {} && \multicolumn{1}{c|} {\checkmark} & \checkmark  &   &\multicolumn{1}{c|} {}  &\checkmark  &  &  \multicolumn{1}{c|} {}& \checkmark  & \tabularnewline

\multicolumn{1}{l|} {Aldoma \etal \cite{zero_order_affordances_detection_2012}} &  & \checkmark  & \multicolumn{1}{c|} {}  & & \multicolumn{1}{c|} {\checkmark} &  & &  \multicolumn{1}{c|} {\checkmark}   & \checkmark  &  & \multicolumn{1}{c|} {} & \checkmark  & \tabularnewline

\multicolumn{1}{l|}{Myers \etal \cite{AffDetectGeometricFeatures_2015}} &  & \checkmark  & \multicolumn{1}{c|} {}  && \multicolumn{1}{c|} {\checkmark} &  &  &\multicolumn{1}{c|} { \checkmark}   &  & \checkmark  &\multicolumn{1}{c|} {}  & \checkmark  & \tabularnewline

\multicolumn{1}{l|}{Herman \etal \cite{Visual_Physical_Affordances_2011}} & \checkmark  &  &  \multicolumn{1}{c|} {} & & \multicolumn{1}{c|} {\checkmark} &  &  & \multicolumn{1}{c|} {\checkmark}  & \checkmark  &  &\multicolumn{1}{c|} {}  & \checkmark  & \tabularnewline

\multicolumn{1}{l|}{Moldovan \etal \cite{Occlusion_RelationalAffordance_2014}} & \checkmark  &  &  \multicolumn{1}{c|} {} & & \multicolumn{1}{c|} {\checkmark} &  & \checkmark  & \multicolumn{1}{c|} {}  & \checkmark  &  & \multicolumn{1}{c|} {} & \checkmark  & \tabularnewline

\multicolumn{1}{l|}{Nguyen et al. \cite{Affordance_Detection_CNN_2016}} &  &  & \multicolumn{1}{c|} { \checkmark}  & \checkmark&  \multicolumn{1}{c|} {}& \checkmark &  & \multicolumn{1}{c|} {\checkmark}   & \checkmark &  &\multicolumn{1}{c|} {}  &   & \checkmark\tabularnewline

\multicolumn{1}{l|}{Sawatzky et al. \cite{Weakly_Supervised_Learning_Affordance_2016}} &  & \checkmark  & \multicolumn{1}{c|} {}  &\checkmark & \multicolumn{1}{c|} {\checkmark}  &  &  & \multicolumn{1}{c|} {\checkmark}   & \checkmark  &  & \multicolumn{1}{c|} {\checkmark}  & \checkmark  & \checkmark \tabularnewline
 
\multicolumn{1}{l|}{Nguyen \etal \cite{Object_based_Affordances_CNN_CRF_2017}}   &  & \checkmark  &  \multicolumn{1}{c|} {} &\checkmark &  \multicolumn{1}{c|} {\checkmark} & \checkmark  &  &  \multicolumn{1}{c|} {} & \checkmark  &  & \multicolumn{1}{c|} {} &  & \checkmark \tabularnewline

\multicolumn{1}{l|}{Do \etal \cite{affordancenet2017_Do}} &  & \checkmark  &  \multicolumn{1}{c|} {}  &\checkmark&   \multicolumn{1}{c|} {} & \checkmark  &  & \multicolumn{1}{c|} {\checkmark}   & \checkmark  &  &\multicolumn{1}{c|} {}  &  & \tabularnewline

\multicolumn{1}{l|}{Chen \etal \cite{DeepDriving_Affordances_2015}} &  & \checkmark  &  \multicolumn{1}{c|} {} &\checkmark &  \multicolumn{1}{c|} {\checkmark} &  & \checkmark  &\multicolumn{1}{c|} {\checkmark}    & \checkmark  &  &\multicolumn{1}{c|} {}   & \checkmark  & \checkmark \tabularnewline
\hline 
\end{tabular*}

\caption{Comparison between affordance detection methods}
\label{table:affordancedetectioncomparison}

\end{table*}

\subsection{ Affordance-Semantic Labeling}
\begin{figure*}
\centering \includegraphics[width=0.85\paperwidth]{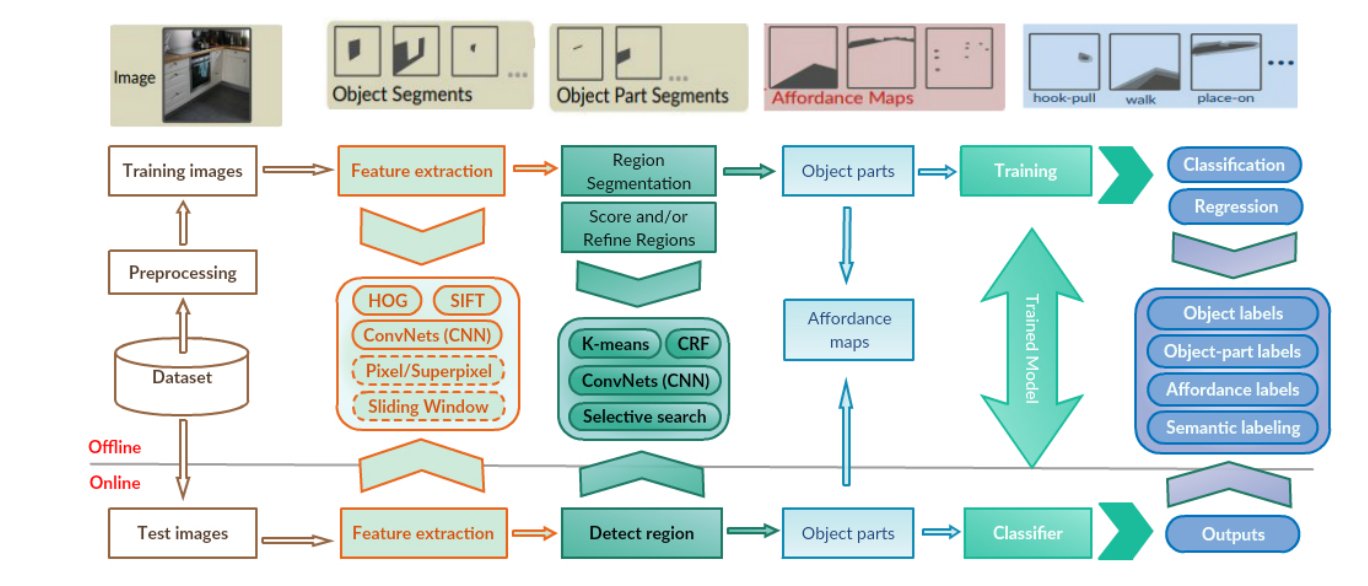}\caption{\label{fig:segmentationdiagram} affordance-based segmentation process diagram.}
\end{figure*}

The affordance semantic labeling task involves assigning pixel-level affordance category labels to relevant regions in an image. This problem combines segmentation and detection tasks and is relatively more challenging. It requires local and global contextual modeling for accurate pixel-level predictions.  Affordance labeling is highly useful for precisely locating where appropriate actions can be performed in a given scene. Figure \ref{fig:segmentationdiagram} shows the most important steps to do segmentation.  

Inspired by the semantic segmentation framework proposed by Eigen and Fergus \cite{predicting_mid_level_cues_2015} for generic object categories, Roy \etal proposed a multi-scale method to segment semantically meaningful affordances  through CNN \cite{multi_scale_affordance_segmentation_2016}. Given a RGB image, their model predicted three types of information: \textbf{1-} the depth map, \textbf{2-} surface normals, \textbf{3-} labels for semantic segmentation. Thereafter, the outputs were merged together in a CNN network to predict the affordance maps. The experiments were performed on the NYU Depth dataset which consists of real-world indoor images \cite{NYU_segmentation_dataset_2012}. This dataset was extended with affordance ground-truth annotations. The authors suggested five affordance labels as shown in Table \ref{table:affordances}.  Although this method introduced a new feature encoding hierarchy based on intermediate semantic segmentation, it does not address the problem of multi-label affordances which is conflicting with the concept of segmentation. This is due to the reason that semantic segmentation aims to assign a single label to all pixels belonging to an object, while each object usually has multiple affordance labels e.g. a knife has both \emph{cut} and \emph{handle} affordances. Likewise, Kim \etal  \cite{Semantic_Labeling_3D_Clouds_2014} performed affordance segmentation by using surface geometry features (e.g. linearity, normal and occupancy) of RGB-D images. The authors suggested six affordances as shown in Table \ref{table:affordances}.

\begin{figure}
\centering \includegraphics[width=0.85\columnwidth,height=0.13\paperheight]{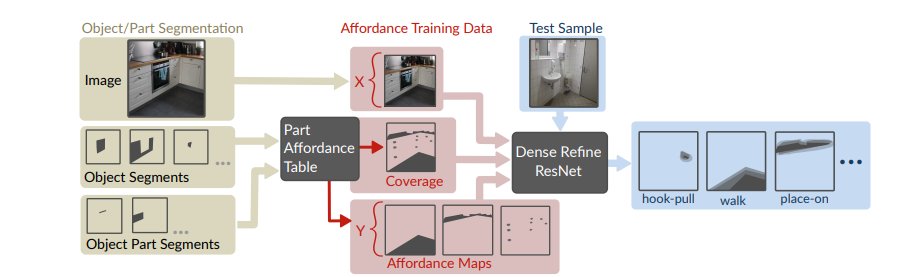}\caption{\label{fig:partssegmentation} The part-segmentation methodology that is proposed by \cite{learning_to_Label_Affordances_2017} They employed segmentation for objects parts and train the the model to predict the affordances. In the same essence, they use manually look up (transfer) table to map between affordances labels and object labels}.
\end{figure}

Most recently, Luddecke and Florentin \cite{learning_to_Label_Affordances_2017} proposed a new method to label affordances in RGB images using a refined version of Residual CNN \cite{Resnet_2016} which was inspired by the work of Piheiro \etal \cite{Refining_ResNet_2016}. As a major novelty, they developed a new cost function to handle multiple affordances in case of incomplete data. In a similar work, Roy and Todorovic \cite{multi_scale_affordance_segmentation_2016} used the concept of action maps which predict the ability of users to do actions at various locations \cite{Action_Maps_2014,Action_Maps_2016}. This results in  pixel-wise affordance segmentations given RGB images. Different from previous works, the authors introduce two original concepts: 
\begin{itemize}\setlength{\topsep}{0em}\setlength{\itemsep}{0em}
\item \textbf{Object Parts} $-$ These are used to detect relevant segments of an object e.g., the surface of a table  is important for placement while table legs are useless as shown in Figure \ref{fig:partssegmentation}. As a consequence, they can train on the only data set that supports object parts i.e. ADE20K \cite{ADE20K_Dataset_Object_Parts_2016}.
\item \textbf{Transfer Table} $-$ It is a manual look-up table to map between object labels and affordance labels. In order to cover the affordance parts, the authors specifically suggested fifteen labels as shown in Table \ref{table:affordances}. \\

\end{itemize}
Apart from that, Table  \ref{table:affordancsemanticlabelingcomparison} compares between the recent methods in the literature.
\begin{table}
\scalebox{0.8}{
\begin{tabular*}{1.25\linewidth}{@{\extracolsep{\fill}}>{\raggedright}m{.26\linewidth}
>{\raggedright}m{0.01\linewidth}
>{\raggedright}m{0.01\linewidth}
>{\raggedright}m{0.01\linewidth}
>{\raggedright}m{0.01\linewidth}
>{\raggedright}m{0.01\linewidth}
>{\raggedright}m{0.01\linewidth}
>{\raggedright}m{0.01\linewidth}
>{\raggedright}m{0.01\linewidth}
>{\raggedright}m{0.01\linewidth}
>{\raggedright}m{0.01\linewidth}
>{\raggedright}m{0.01\linewidth}
>{\raggedright}m{0.01\linewidth}
>{\raggedright}m{0.01\linewidth}
}
\toprule 
\multirow{2}{*}{} & \multicolumn{2}{|l|}{Input } &  \multicolumn{2}{l|}{Features}   & \multicolumn{2}{l|}{Evaluation} & \multicolumn{3}{l|}{Training} & \multicolumn{2}{l}{Model}\tabularnewline
 \midrule 
 &  \multicolumn{1}{|c}{\rotatebox[origin=c]{90}{2D}}  &   \rotatebox[origin=c]{90}{3D}  &   \multicolumn{1}{|c}{\rotatebox[origin=c]{90}{Feature learning}} &   \rotatebox[origin=c]{90}{Hand crafted} &   \multicolumn{1}{|c}{\rotatebox[origin=c]{90}{Real robot}}  &   \rotatebox[origin=c]{90}{Benchmark} &   \multicolumn{1}{|c}{\rotatebox[origin=c]{90}{Supervised}}  &   \rotatebox[origin=c]{90}{Unsupervised}  &   \rotatebox[origin=c]{90}{Weakly Supervised} &   \multicolumn{1}{|c}{\rotatebox[origin=c]{90}{Mathematical}} &     \multicolumn{1}{l}{\rotatebox[origin=c]{90}{Neural}}\tabularnewline
 \midrule 

 Roy \etal \cite{multi_scale_affordance_segmentation_2016} & \multicolumn{1}{|c}{\checkmark} &  \multicolumn{1}{c|}{} &    & \multicolumn{1}{c|}{\checkmark} &    & \multicolumn{1}{c|}{\checkmark} & \checkmark &  &\multicolumn{1}{c|}{}  &\checkmark & \multicolumn{1}{l}{}\tabularnewline

\rule{0pt}{3ex}Kim \etal \cite{Semantic_Labeling_3D_Clouds_2014} & \multicolumn{1}{|c}{}  &   \multicolumn{1}{c|}{\checkmark } &  & \multicolumn{1}{c|}{\checkmark} & \checkmark   &  \multicolumn{1}{c|}{}& \checkmark &  &\multicolumn{1}{c|}{}  & \checkmark &\multicolumn{1}{l}{} \tabularnewline

\rule{0pt}{3ex}Luddecke \etal \cite{learning_to_Label_Affordances_2017}&\multicolumn{1}{|c}{ \checkmark} &  \multicolumn{1}{c|}{} &   \checkmark &   \multicolumn{1}{c|}{}&   & \multicolumn{1}{c|}{\checkmark} & \checkmark &  &\multicolumn{1}{c|}{}  &  & \multicolumn{1}{l}{\checkmark}\tabularnewline

\rule{0pt}{3ex} Nguyen \etal \cite{Object_based_Affordances_CNN_CRF_2017}
&\multicolumn{1}{|c}{ } 
&  \multicolumn{1}{c|}{\checkmark} 
&   \checkmark 
&   \multicolumn{1}{c|}{}
&  
& \multicolumn{1}{c|}{\checkmark} 
& \checkmark
& 
&\multicolumn{1}{c|}{}  
& 
& \multicolumn{1}{l}{\checkmark}\tabularnewline

\rule{0pt}{3ex} Do \etal \cite{affordancenet2017_Do} &\multicolumn{1}{|c}{ }
&  \multicolumn{1}{c|}{\checkmark}
&   \checkmark 
&   \multicolumn{1}{c|}{}
&\checkmark
& \multicolumn{1}{c|}{\checkmark} 
& \checkmark 
&  
&\multicolumn{1}{c|}{}  
&
& \multicolumn{1}{l}{\checkmark}\tabularnewline

\bottomrule
\end{tabular*}}
\caption{Comparison between affordance-semantic labeling methods}
\label{table:affordancsemanticlabelingcomparison}
\end{table}

\subsection{Affordance as a Context }
Affordances are inter-linked with both physical and semantic characteristics of an object and a scene. Object affordances can provide useful clues about object properties such as their category, location and function. In this section, we describe research efforts that aim to use affordance relationships as a context for other associated tasks such as action recognition, object detection and gesture recognition. 

In an early work, Fitzpatrick \etal \cite{Learning_Affordances_2003} proposed a method that allows a robot to learn how to segment the objects through imitation. The goal was to allow the humanoid robots  to understand through acting. Similarly, through imitating the human actions on objects, a robot can learn object affordances e.g., whether a spherical shape is rollable or a cubic shape is slide-able.  In addition to predicting affordances, it could interpret other's actions. The intertwining of objects and actions, which is formally known by motor actions  \cite{Learning_Imitation_2002}, to learn about objects affordances or segment objects was a promising at that time. Montesano \etal \cite{Learning_Affordances_Sensory_Motor_2008} sought to learn affordances through robot-environment interactions. Additionally, they used affordance labels such as eatable, movable and graspable as sensing capabilities. Local features e.g., color characteristics were used to detect the shape of the object which was eventually used to detect these affordances. Then the robot learned the model of grasping and used affordances through imitation and self observation. In the same context of motor actions, They proposed a probabilistic model based on Markov Chain Monte Carlo sampling. Inspired by Montesano \etal\cite{Modeling_Affordances_Bayesian_2007}, Lopes \etal \cite{Affordance_Learning_Imitation_2007} proposed a probabilistic technique based on Bayesian Networks to learn affordances, and thereafter these learned affordances were used to recognize the demonstrations of an agent and learn the given task. Based on the affordances (e.g. tappable or graspable) and through self observation, the robot could relate the action with the resulting effects. However, learning affordances was employed for only single objects. Likewise, Ugur \etal \cite{Behavioural_Affordance_2011} used self-interaction and self-observation to build their model. However, they provided behavioral parameters to enhance the accuracy. In contrast to previous methods \cite{Affordance_Learning_Imitation_2007,Learning_Affordances_Sensory_Motor_2008}, the authors used unsupervised clustering to segment grasping through 300 trials whereas SVM was used to learn affordance labels. Varadarajan \etal \cite{AfNet_2012,Afrob_2012} developed a dataset to build knowledge ontologies similar to MIT ConceptNet \cite{ConceptNet_2007} and KnowRob Semantic Map \cite{KnowRob_2010}, but for household RGB-D images. They presented various affordance features such as grasp, material and structural. The authors built affordance filtrations which started by localizing the affordances, then they identified the entities related  to that object and named it affordance duals. After that, they looked for all the entities that share the same affordance with that object. Moreover, they used a semantic part segmentation algorithm \cite{Object_Part_Segment_2011} for segmentation whereas modified the Levelberg Marquardt Algorithm (LMA) \cite{Levenberg_Marquardt_algorithm_1978} using swarm PSO in order to recognize the objects. 

 Gupta et al. \cite{Human_workspace_affordance_2011} modeled
affordances in 3D indoor images to detect the workspace based on human
poses while they used upright, lay down, reach sit as the human poses.
Inspired by the idea of Gibson which says that the recognition of
objects based on the function is better than the visual appearance \cite{GibsonAffordanceWord_1977},
Castellini et al. \cite{Affordanceto_Improve_Recognition_2011} proposed
using affordances as visual features and motor features, which are defined by kinematic features of the hand when grasping (e.g. time and instance of contact), to enhance the accuracy of recognition. Additionally, these features were defined as human-hand poses while grasping an object. The authors introduced the CONTACT VMGdb dataset which has visual features and kinematics content of grasping in various illumination conditions. They focused their learning on five affordance labels (cylindric power, flat, pinch, spherical and tripodal) along with various objects for grasping.
Moldovan \etal\cite{Spatial_Relational_Affordances_2018} extended the model of \cite{Learning_Affordance_by_Imitation, Learning_Affordances_Sensory_Motor_2008} that considers three related concepts: actions, objects and effects as shown in figure \ref{fig:objects_actions_Image}. They used spatial relations, which were defined by distance between objects and affordances to tackle the problem of multiple-objects manipulation. They used probabilistic programming with logical rules and probabilities to build their model. On the contrary, Lopes et al. \cite{Learning_Affordance_by_Imitation} used object affordances as priori information to enhance gesture recognition and reduce ambiguities based on motor terms. 
\begin{figure}
\centering \includegraphics[width=\columnwidth,height=0.15\paperheight]{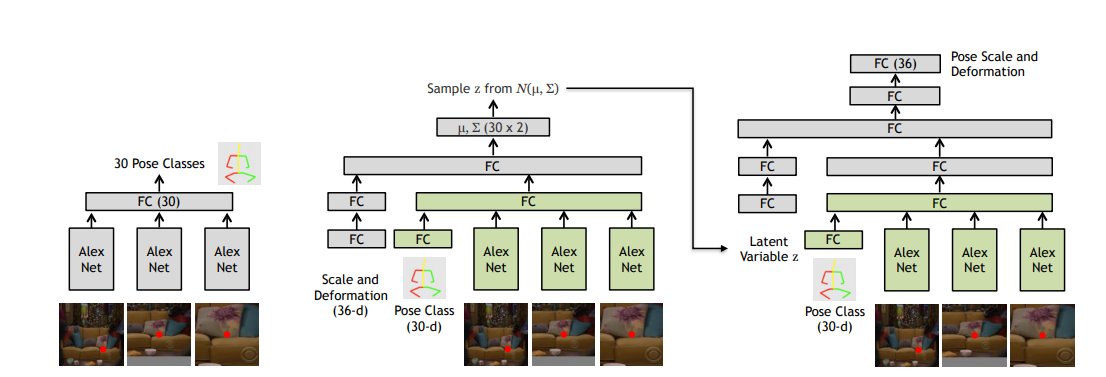}\caption{\label{fig:sitcom} The deep learning architecture that is proposed by \cite{BingeWang_2017_CVPR}. The three parts of images from left to right as following: 1- Classification network 2- Variational AutoEncoder (VAE) 3- VAE decoder. VAE encoders and decoders share the weights in layers which have green color}.
\end{figure}
Recently, Wang et al. \cite{BingeWang_2017_CVPR} used Variational-Auto Encoders (VAE) \cite{Variational_Auto_Encoder_2013} to build their model to predict the affordance poses. Based on the location, the algorithm classifies it into one of the 30 pose classes. Thereafter, it uses a VAE to extract the deformation of this pose. They proved that the nonexistent poses can be predicted using ConvNets and VAE. In addition, the authors claimed that deep learning based efforts are quite few in this area due to the non-existence of a big dataset for learning affordances. For this reason, they build a large-scale dataset from sitcoms to learn affordances.

\begin{figure}
\centering
 \includegraphics[width=0.8\columnwidth]{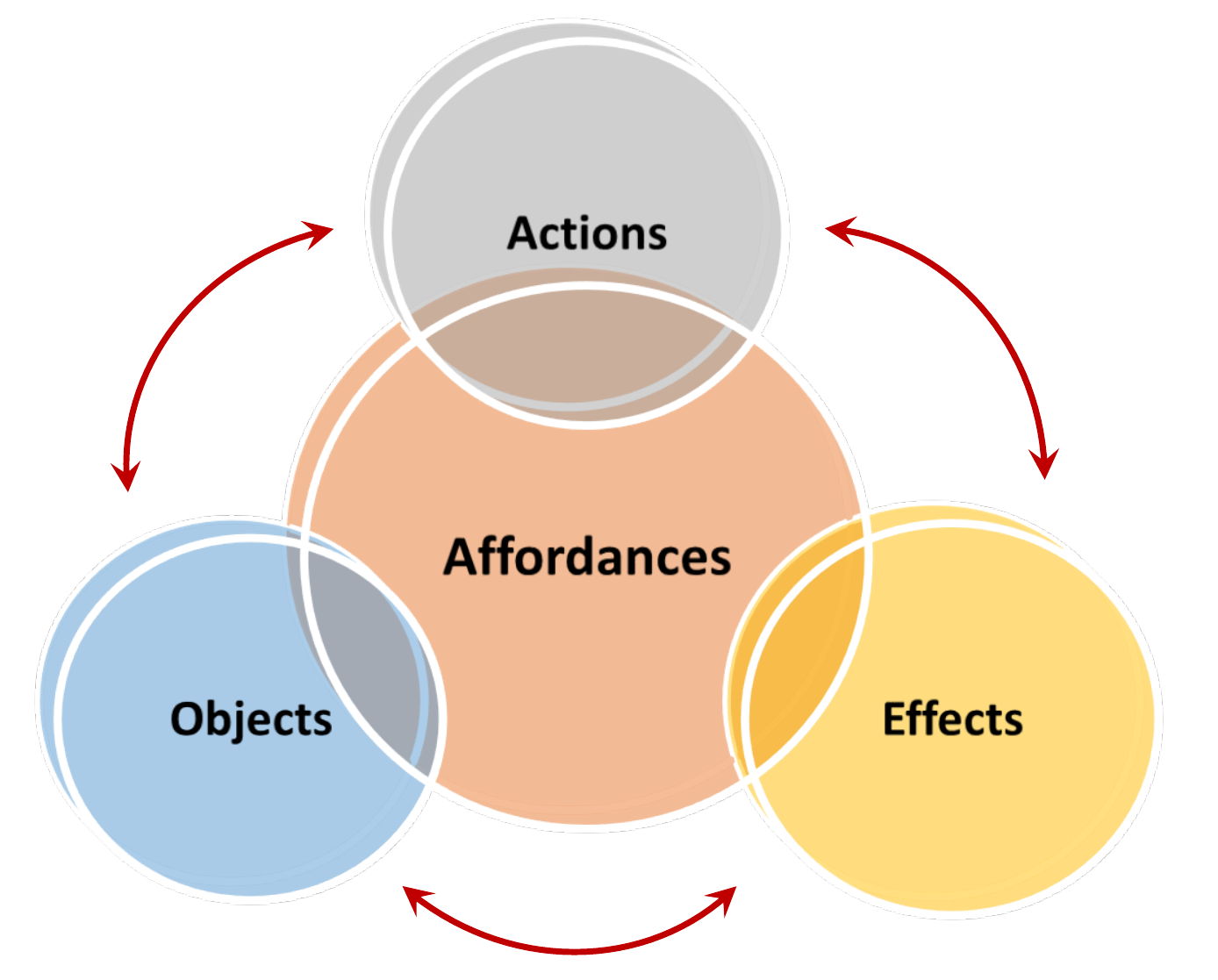}\caption{ The relations between objects, actions and effects }.\label{fig:objects_actions_Image}
\end{figure}

Sun \etal \cite{object_object_interaction_2014} presented a method to learn affordances from object's interactions. Given a video with labeled sequenced frames, their approach produces interactive motion models between pairs of objects and then represents it in a Bayesian network with human actions. Ruiz and Cuevas \cite{Geometric_properties_bisector_2000, Interaction_Bisector_Surface_2014} proposed using bisector surfaces and developed them to include rational weighting, provenance vectors and affordance key points. They aimed to study the affordance locations for objects over one another (e.g. man riding bike, placing pot in the kitchen shelf or where the kids ride bike in a flat) and simultaneously detect unseen views of the object.
Thermos \etal \cite{Affordance_Sensorimotor_Recognition_2017} presented a deep learning paradigm to investigate the problem of sensorimotor 3D object recognition. They employed a biological neural network architecture (VGG-16 network \cite{VGG16_simonyan2014}) to fuse multiple evidence sources to learn affordances. They used fifteen affordance types as shown in Table \ref{table:affordances}. Some of these types describe complex affordances like "squeeze'' and continuous like "write''. Moreover, they introduced a new RGB-D dataset  which has human-interaction and affordance types to test their method. All of the above mentioned methods have been summarized in Table \ref{table:affordancemethodscomparison} to show what has been done and what should be solved. 

\begin{table}[H]
\scalebox{0.82}{
\begin{tabular*}{1.2\linewidth}{@{\extracolsep{\fill}}>{\raggedright}m{.25\linewidth}
>{\raggedright}m{0.01\linewidth}
>{\raggedright}m{0.01\linewidth}
>{\raggedright}m{0.01\linewidth}
>{\raggedright}m{0.01\linewidth}
>{\raggedright}m{0.01\linewidth}
>{\raggedright}m{0.01\linewidth}
>{\raggedright}m{0.01\linewidth}
>{\raggedright}m{0.01\linewidth}
>{\raggedright}m{0.01\linewidth}
>{\raggedright}m{0.01\linewidth}
>{\raggedright}m{0.01\linewidth}
>{\raggedright}m{0.01\linewidth}
>{\raggedright}m{0.01\linewidth}
}

 \cline{1-12} 
 \rule{0pt}{3ex} 
\multirow{2}{*}{} & \multicolumn{2}{|l|}{Input } &  \multicolumn{2}{l|}{Features}   & \multicolumn{2}{l|}{Evaluation} & \multicolumn{3}{l|}{Training} & \multicolumn{2}{l}{Model}\tabularnewline

 \cline{2-12} 
 \rule{0pt}{3ex} 
 &  \multicolumn{1}{|c}{\rotatebox[origin=c]{90}{2D}}  &   \multicolumn{1}{c|}{\rotatebox[origin=c]{90}{3D}}  &   \multicolumn{1}{c}{\rotatebox[origin=c]{90}{   Feature learning   }} &   \multicolumn{1}{c|}{\rotatebox[origin=c]{90}{Handcrafted}} &   \multicolumn{1}{c}{\rotatebox[origin=c]{90}{Real Robot}}  &   \multicolumn{1}{c|}{\rotatebox[origin=c]{90}{Benchmark}} &   \multicolumn{1}{c}{\rotatebox[origin=c]{90}{Supervised}}  &   \rotatebox[origin=c]{90}{Unsupervised}  &   \multicolumn{1}{c|}{\rotatebox[origin=c]{90}{  Semi-supervised  }} &   \multicolumn{1}{c}{\rotatebox[origin=c]{90}{Mathematical}} &     \multicolumn{1}{l}{\rotatebox[origin=c]{90}{Neural}}\tabularnewline

 \cline{1-12} 
\rule{0pt}{3ex} 
 Fitzpatrick \etal \cite{Learning_Affordances_2003}  & \multicolumn{1}{|c}{\checkmark} &   \multicolumn{1}{c|}{}  &  & \multicolumn{1}{c|}{\checkmark} & \checkmark   & \multicolumn{1}{c|}{} &  &  \checkmark&\multicolumn{1}{c|}{}  &\checkmark & \multicolumn{1}{l}{}\tabularnewline

 Lopes \etal \cite{Affordance_Learning_Imitation_2007} & \multicolumn{1}{|c}{\checkmark}  &   \multicolumn{1}{c|}{} &  &  \multicolumn{1}{c|}{\checkmark} & \checkmark   &  \multicolumn{1}{c|}{}&  &  &\multicolumn{1}{c|}{\checkmark}  & \checkmark &\multicolumn{1}{l}{} \tabularnewline

 Montesano \etal \cite{Learning_Affordances_Sensory_Motor_2008} &\multicolumn{1}{|c}{ \checkmark} & \multicolumn{1}{c|}{}   &  &   \multicolumn{1}{c|}{\checkmark}&\checkmark   & \multicolumn{1}{c|}{} &  &\checkmark  &\multicolumn{1}{c|}{}  & \checkmark & \multicolumn{1}{l}{}\tabularnewline

 Ugur \etal \cite{Behavioural_Affordance_2011}&\multicolumn{1}{|c}{ \checkmark} &  \multicolumn{1}{c|}{}   &  &   \multicolumn{1}{c|}{\checkmark}&\checkmark   & \multicolumn{1}{c|}{} &  & \checkmark &\multicolumn{1}{c|}{}  & \checkmark & \multicolumn{1}{l}{}\tabularnewline

 Varadarajan \etal \cite{AfNet_2012,Afrob_2012} &\multicolumn{1}{|c}{ } &  \multicolumn{1}{c|}{\checkmark}   &  &   \multicolumn{1}{c|}{\checkmark}&   & \multicolumn{1}{c|}{\checkmark} & \checkmark &  &\multicolumn{1}{c|}{}  & \checkmark & \multicolumn{1}{l}{}\tabularnewline

 Gupta \etal \cite{Human_workspace_affordance_2011}  &\multicolumn{1}{|c}{ } & \multicolumn{1}{c|}{ \checkmark}   &  &   \multicolumn{1}{c|}{\checkmark}&   & \multicolumn{1}{c|}{\checkmark} &  \checkmark &  &\multicolumn{1}{c|}{}  & \checkmark & \multicolumn{1}{l}{}\tabularnewline

Castellini \etal \cite{Affordanceto_Improve_Recognition_2011} &\multicolumn{1}{|c}{ } &  \multicolumn{1}{c|}{\checkmark}   &  &   \multicolumn{1}{c|}{\checkmark}&   & \multicolumn{1}{c|}{\checkmark} & \checkmark &  &\multicolumn{1}{c|}{}  & \checkmark & \multicolumn{1}{l}{}\tabularnewline

Grabner \etal  \cite{WhatMakesChairAChair_2011}&\multicolumn{1}{|c}{ } &  \multicolumn{1}{c|}{\checkmark}   & &   \multicolumn{1}{c|}{\checkmark}&   & \multicolumn{1}{c|}{\checkmark} & \checkmark &  &\multicolumn{1}{c|}{}  &\checkmark  & \multicolumn{1}{l}{}\tabularnewline

 Lopes \etal \cite{Learning_Affordance_by_Imitation} &\multicolumn{1}{|c}{ \checkmark} &  \multicolumn{1}{c|}{}   &  &   \multicolumn{1}{c|}{\checkmark}&\checkmark   & \multicolumn{1}{c|}{} &  &  &\multicolumn{1}{c|}{\checkmark}  &\checkmark	  & \multicolumn{1}{l}{}\tabularnewline

Wang \etal \cite{BingeWang_2017_CVPR} &\multicolumn{1}{|c}{ \checkmark} &  \multicolumn{1}{c|}{}   & \checkmark &   \multicolumn{1}{c|}{}&  & \multicolumn{1}{c|}{\checkmark} & & \checkmark  &\multicolumn{1}{c|}{}  &  & \multicolumn{1}{l}{\checkmark}\tabularnewline

Sun \etal \cite{object_object_interaction_2014} &\multicolumn{1}{|c}{ \checkmark} &  \multicolumn{1}{c|}{}   &  &   \multicolumn{1}{c|}{\checkmark}&   & \multicolumn{1}{c|}{\checkmark} & \checkmark &  &\multicolumn{1}{c|}{}  &  \checkmark & \multicolumn{1}{l}{}\tabularnewline

Thermos \etal \cite{Affordance_Sensorimotor_Recognition_2017} &\multicolumn{1}{|c}{ } & \multicolumn{1}{c|}{\checkmark }   & \checkmark &   \multicolumn{1}{c|}{}&   & \multicolumn{1}{c|}{\checkmark} & \checkmark &  &\multicolumn{1}{c|}{}  &  & \multicolumn{1}{l}{\checkmark}\tabularnewline

\bottomrule
\end{tabular*}}
\caption{Comparison between contextual affordance recognition methods}
\label{table:affordancemethodscomparison}
\end{table}

\begin{figure*}
\centering \includegraphics[width=\textwidth]{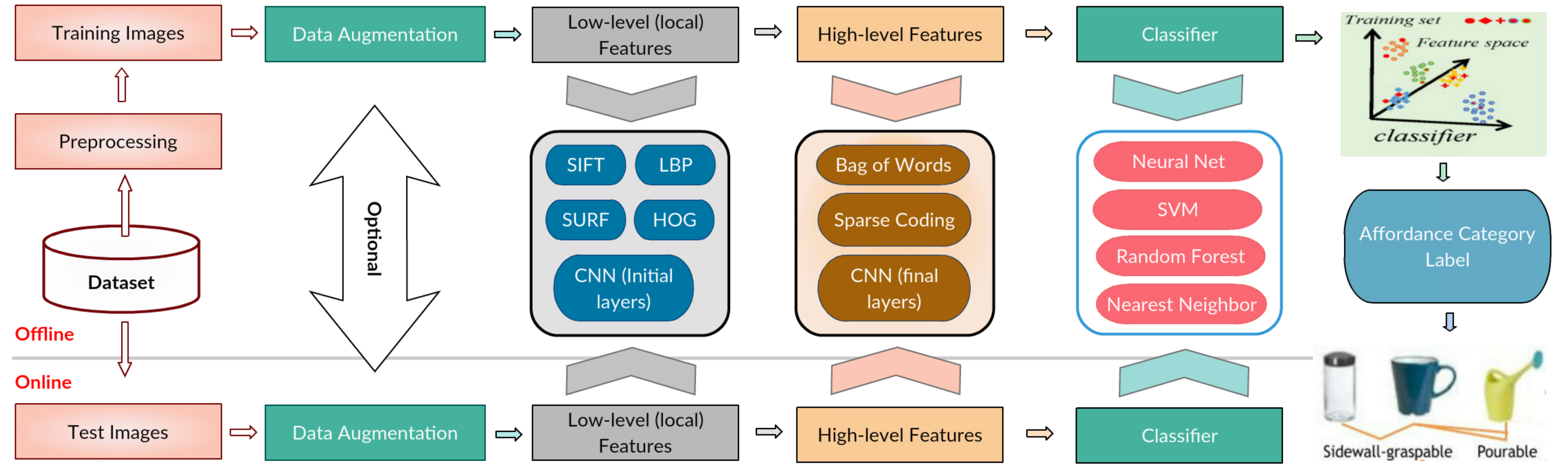}\caption{\label{fig:Categorization} Affordance classification process through machine learning techniques. }
\end{figure*}

\subsection{Affordance Categorization}

The affordance categorization task aims to tag an image with the relevant set of affordance labels. To this end, a general approach is to represent an image in the form of discriminative features and employ a classifier to assign affordance labels. This task is relatively simple compared to affordance detection and segmentation, which also localize affordance categories (see Figure \ref{fig:classificationFlowchart}).

Varadarajan and Vincze \cite{Parallel_DL_Affordance_Categories_2013} proposed hybrid parallel architecture of deep learning and suggestive activation (PDLSA) to overcome the problems of deep learning like uni-modality and serialization in order to categorize the objects based on the affordance features. The authors extracted the semantic features of affordances as proposed in \cite{Afrob_2012} as well as the structural and material features to enhance the recognition results. The Washington RGB-D dataset was used to test the efficacy of their model. Sun \etal employed object categorization as an intermediate step to infer the affordances more correctly \cite{Learning_Affordances_Category_2010}. They developed a visual category-based affordance model, which encoded the relationships among visual features, learned categories and affordances in probabilistic form. Such a probabilistic modeling allows knowledge transfer and enhance accuracy, especially when limited annotated data is available. Additionally, they addressed the problem of incremental learning of affordances. The authors suggested seven object categories and six affordance labels as shown in Table \ref{table:affordances}. However, this study has been devoted to indoor buildings and it was not enough to treat all the cases such as defining that the table is movable requires knowing its physical attributes. In \cite{WhatMakesChairAChair_2011}, the authors have categorized the images based on functions to enhance
the performance of detection.
The concept of bootstrapping, which uses the past knowledge to accelerate the learning process, has been applied to affordance learning \cite{Bootstrapping_semantics_tools_2016,Bootstrapping_Paired_objects_2014}. Schoeler \etal \cite{Bootstrapping_semantics_tools_2016} proposed to infer any possible usage of a tool even if that usage is possible through another main tool e.g., using stone as hammer or using helmet instead of water cup. They sought to divide the tools into six functional categories (contain, cut, hit, hook, poke and sieve). Afterwards, an ontology of tool functions was created to allow a deeper understanding  by exploring their usage in absence of main tool. The authors developed their main algorithm in three steps: 
\textbf{(1)} Part Segmentation through Constrained Planar Cuts (CPC) algorithm \cite{segmentation_Constrained_planar_cuts_Part_2015},
\textbf{(2)} Extraction of part-based visual features via a Signature of Histograms of Orientations (SHOT) descriptor
\cite{signature_histograms_of_orientations_SHOT_2010,ViewPoint_Invariant_shape_2013},
\textbf{(3)} The pose of individual parts with respect to each other is encoded via a Pose Signature that models the alignment and attachment between parts. Although the authors introduced a new idea that was thoroughly tested on a 3D synthetic dataset. However, their method gets confused when many tools exist to perform the same action. Also, the selected tool for some action may not be well-suited due to its size or shape. To avoid these problems they need to find the best correspondences between tools and objects.

Given basic affordances, Ugur \etal \cite{Bootstrapping_Paired_objects_2014}  proposed a bootstrapping method to learn the complex affordances through the relational affordances or so-called paired-object affordances. They evaluated their approach using a real robot on various object shapes  like boxes, spheres and cylinders. Additionally they trained the robot to perform actions such as side-poke and stack. Similarly, Fichtl \etal \cite{Bootstrapping_semantics_tools_2016} addressed the problem of "related affordances", which denotes the cases where affordances are related together e.g. open kitchen door to fetch the glass from inside. They used pose and size as the visual features to build their model.
Abelha \etal \cite{Substitue_Tools_2016,Substitute_Tools_fromWeb_2017,Transfer_tools_Affordances_2017} proposed methods to identify tools and their substitutes based on matching superquadrics \cite{superquadrics_2013} of these tools in point cloud data. Mar \etal \cite{Affordance_Learning_multimodal_2015, SOM_2017} used learning by exploration as the methodology to train the iCub robot. In the latter work \cite{SOM_2017}, they categorized the objects before using parallel Self Organizing Maps (SOM). Following their
study \cite{Function_Object_Activity_decriptor_2013}, Pieropan \etal used the same method to learn affordances rather than functionalities \cite{Function_descriptor_Object_Affordance_2015}. Kjellstr{\"o}m \etal \cite{Action_Recognition_Affordance_2011} proposed functional categorization to learn object affordances through human demonstration. They used a Conditional Random Field (CRF) and factorial conditional random field to train the model and infer the object's affordances and actions even when they are novel. Overall, Table \ref{table:affordanceclassificationcomparison} presents detailed comparison between affordance classification methods. 

\begin{table*}
\centering
\scalebox{0.9}{
\begin{tabular*}{1\linewidth}{@{\extracolsep{\fill}}>{\raggedright}m{.22\linewidth}
>{\raggedright}m{0.01\linewidth}
>{\raggedright}m{0.01\linewidth}
>{\raggedright}m{0.01\linewidth}
>{\raggedright}m{0.01\linewidth}
>{\raggedright}m{0.01\linewidth}
>{\raggedright}m{0.01\linewidth}
>{\raggedright}m{0.01\linewidth}
>{\raggedright}m{0.01\linewidth}
>{\raggedright}m{0.01\linewidth}
>{\raggedright}m{0.01\linewidth}
>{\raggedright}m{0.01\linewidth}
>{\raggedright}m{0.01\linewidth}
>{\raggedright}m{0.01\linewidth}
>{\raggedright}m{0.01\linewidth}
}
\toprule 
\multirow{2}{*}{} & \multicolumn{3}{|l|}{Input } &  \multicolumn{2}{l|}{Features}   & \multicolumn{3}{l|}{Evaluation} & \multicolumn{4}{l|}{Training} & \multicolumn{2}{l}{Abstraction}\tabularnewline
  
 \cline{2-15} 
 &  \multicolumn{1}{|c}{\rotatebox[origin=c]{90}{2D}}  &   \rotatebox[origin=c]{90}{3D} &    \multicolumn{1}{c|}{\rotatebox[origin=c]{90}{Multimodal}}   &  \rotatebox[origin=c]{90}{Feature learning} &   \multicolumn{1}{c|}{\rotatebox[origin=c]{90}{Handcrafted}} &   \rotatebox[origin=c]{90}{Real Robot}  & \rotatebox[origin=c]{90}{Simulation}&   \multicolumn{1}{c|}{\rotatebox[origin=c]{90}{Benchmark}} &   \rotatebox[origin=c]{90}{Supervised}  &   \rotatebox[origin=c]{90}{Unsupervised} & \rotatebox[origin=c]{90}{Self-supervised} &    \multicolumn{1}{c|}{\rotatebox[origin=c]{90}{Semi-supervised}} &   \rotatebox[origin=c]{90}{Mathematical} &     \multicolumn{1}{l}{\rotatebox[origin=c]{90}{Neural}}\tabularnewline
 \midrule 

 Varadarajan \etal\cite{Parallel_DL_Affordance_Categories_2013}  & \multicolumn{1}{|c}{} &  \checkmark& \multicolumn{1}{c|}{}  & \checkmark & \multicolumn{1}{c|}{} &   & & \multicolumn{1}{c|}{\checkmark } & \checkmark& &  &\multicolumn{1}{c|}{}  & & \multicolumn{1}{l}{\checkmark}\tabularnewline
  
 Sun et al. \cite{Learning_Affordances_Category_2010} & \multicolumn{1}{|c}{\checkmark}  & &   \multicolumn{1}{c|}{} &  &  \multicolumn{1}{c|}{\checkmark} &  \checkmark & &  \multicolumn{1}{c|}{}& \checkmark  &  & &\multicolumn{1}{c|}{}  & \checkmark &\multicolumn{1}{l}{} \tabularnewline
 
 Schoeler \etal \cite{Bootstrapping_semantics_tools_2016}  &\multicolumn{1}{|c}{ } & \checkmark&\multicolumn{1}{c|}{ }   &  &   \multicolumn{1}{c|}{\checkmark}&&   & \multicolumn{1}{c|}{\checkmark} &  & \checkmark & &\multicolumn{1}{c|}{}  & \checkmark & \multicolumn{1}{l}{}\tabularnewline

 Ugur \etal \cite{Bootstrapping_Paired_objects_2014}&\multicolumn{1}{|c}{} &  \checkmark&\multicolumn{1}{c|}{}   &  &   \multicolumn{1}{c|}{\checkmark}&\checkmark   & & \multicolumn{1}{c|}{} &  &  \checkmark &&\multicolumn{1}{c|}{}  & \checkmark & \multicolumn{1}{l}{}\tabularnewline

 Fichtl \etal \cite{Bootstrapping_semantics_tools_2016} &\multicolumn{1}{|c}{ } &  \checkmark&\multicolumn{1}{c|}{}   &  &   \multicolumn{1}{c|}{\checkmark}&  &\checkmark & \multicolumn{1}{c|}{}& & &  &\multicolumn{1}{c|}{\checkmark}  & \checkmark & \multicolumn{1}{l}{}\tabularnewline

  Abelha \etal\cite{Substitue_Tools_2016,Substitute_Tools_fromWeb_2017,Transfer_tools_Affordances_2017} &\multicolumn{1}{|c}{ } &  \checkmark&\multicolumn{1}{c|}{}   &  &   \multicolumn{1}{c|}{\checkmark}& &  & \multicolumn{1}{c|}{\checkmark} & \checkmark & & &\multicolumn{1}{c|}{}  & \checkmark & \multicolumn{1}{l}{}\tabularnewline

Mar \etal \cite{Affordance_Learning_multimodal_2015}&\multicolumn{1}{|c}{ } &  &\multicolumn{1}{c|}{\checkmark}   &  &   \multicolumn{1}{c|}{\checkmark}& \checkmark&  & \multicolumn{1}{c|}{} &   & \checkmark  & &\multicolumn{1}{c|}{}  & \checkmark & \multicolumn{1}{l}{}\tabularnewline

Mar \etal \cite{SOM_2017}&\multicolumn{1}{|c}{ } &  \checkmark&\multicolumn{1}{c|}{}   & &   \multicolumn{1}{c|}{\checkmark}& & \checkmark & \multicolumn{1}{c|}{} &  & &\checkmark  &\multicolumn{1}{c|}{}  &  & \multicolumn{1}{l}{\checkmark}\tabularnewline

  Pieropan \etal \cite{Function_Object_Activity_decriptor_2013} &\multicolumn{1}{|c}{ } & \checkmark& \multicolumn{1}{c|}{}   &  &   \multicolumn{1}{c|}{\checkmark}& &  & \multicolumn{1}{c|}{\checkmark} &  &  \checkmark &&\multicolumn{1}{c|}{}  &\checkmark	  & \multicolumn{1}{l}{}\tabularnewline

Kjellstr{\"o}m \etal \cite{Action_Recognition_Affordance_2011} &\multicolumn{1}{|c}{ \checkmark} &&  \multicolumn{1}{c|}{}   & &   \multicolumn{1}{c|}{ \checkmark}& & & \multicolumn{1}{c|}{\checkmark} & \checkmark& &  &\multicolumn{1}{c|}{}  &  \checkmark & \multicolumn{1}{l}{}\tabularnewline

\bottomrule
\end{tabular*}}
\caption{Comparison between affordance classification methods.}
\label{table:affordanceclassificationcomparison}
\end{table*}

\subsection{Affordance-based activity recognition}

\begin{figure*}
\centering \includegraphics[width=\textwidth]{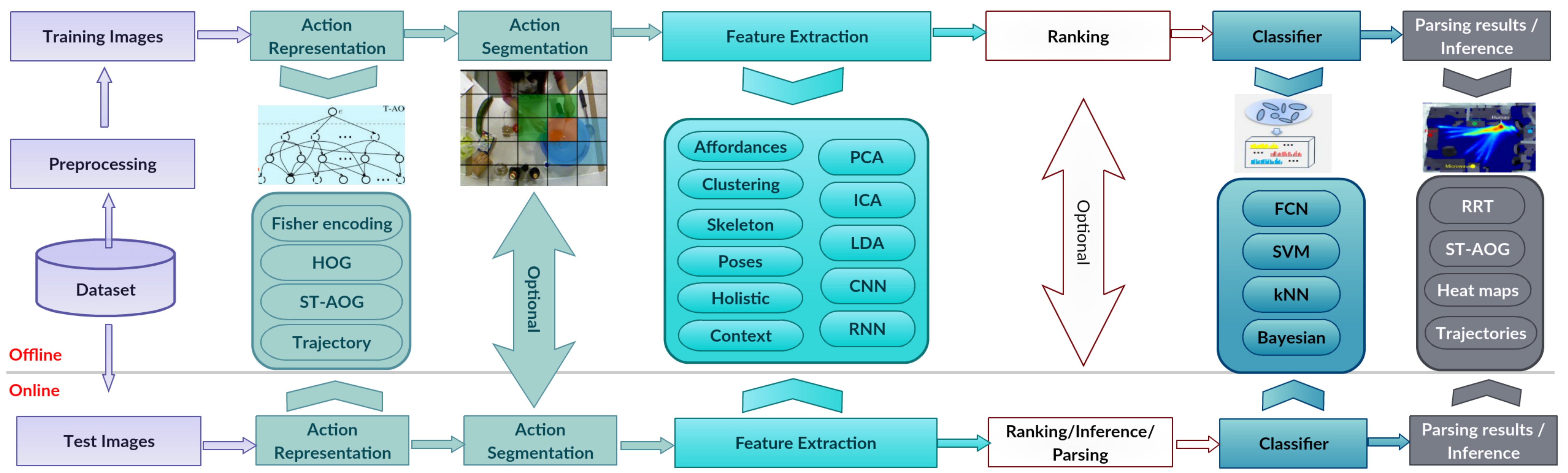}\caption{\label{fig:activityrecognition} Affordance-based activity recognition process. }.
\end{figure*}

Enabling seamless human-robot interaction is a crucial step towards ubiquitous use of personal robots. It is a multidisciplinary research area which overlaps with Robotics, Human-Computer Interaction, Cognitive Science, Artificial Intelligence, Action Recognition and Affordance Prediction. To understand the interaction between robot and its environment, the affordance of objects should be predicted which can be a very useful cue for activity recognition as shown in Figure \ref{fig:activityrecognition}. 

Due to the close relationship between affordances, actions and humans, its use in recognizing human action and hence activities has been investigated. In this section, the relation between affordances and actions will be reviewed. Koppula \etal \cite{Human_Activities_Affordance_2013} learned the human activities from RGB-D videos considering object affordances. They structured their scheme using a graphical representation where the nodes represented the sub-activities and objects; and the edges referred to the object's affordances and the relations between human and objects. Additionally, they learned their model through a structural support vector machine algorithm. In a recent study, Koppula \etal \cite{Anticipating_human_activities_affordances_2016} presented a modified version of their previous work \cite{Human_Activities_Affordance_2013} where they used a Conditional Random Field (CRF) to represent the model. They merged CRF structure with object affordances and sub-activities to form so-called Anticipatory Temporal CRF (ATCRF). Following these studies \cite{Human_Activities_Affordance_2013,Anticipating_human_activities_affordances_2016}, Jain \etal merged spatial-temporal graph with a Recurrent Neural Network (RNN) to address problems in graphical models \cite{Structura_RNN_2016}. They trained and tested different kinds of spatial-temporal cases like motion modeling, human activity prediction and action anticipation. 
Qi \etal \cite{Human_Activities_Grammar_2017} represented affordances, human actions and interacting objects in a Spatial-Temporal And-Or graph (ST-AOG) to predict human activities in RGB-D videos. They built the model in two main stages: video parsing and activity prediction. The parsing is done using segmentation by a dynamic programming approach and later label refinement using Gibbs sampling. For activity prediction, it depended on an Earley parser \cite{Earley_Parser_1970} to predict sub-activites and all the learned cues (parsed graph and sub-activities) to estimate human activity. Vu \etal \cite{Predicting_actions_static_scenes_2014} challenged that various scenes under the same category have similar functional features. They described scenes in terms of functionalities to predict actions from static images. Dutta and Zielinska \cite{Action_prediction_2017} presented a novel method to predict next action based on object affordances and human interaction. They represented the model in a spatio-temporal based probabilistic state automaton. The generated motion trajectory was used to build action heat maps that led to infer next actions. Shu et al. \cite{HRI_SocialAffordances_2016} proposed learning social affordances from human to human interactions. They represented their model in a graphical scheme that has nodes as subevents/subgoals. They provided a RGB-D video dataset (HHOI) to describe human to human interactions. Given a RGB-D video, Shu \etal \cite{social_affordance_grammar_2017} learned social affordance grammars and then represented them as a ST-AOG to perform motion modeling. To sum up, Table \ref{table:affordancebasedactivityrecognitioncomparison} compares between affordance-based activity recognition techniques.

\begin{figure}
\centering \includegraphics[width=0.85\columnwidth,height=0.27\paperheight]{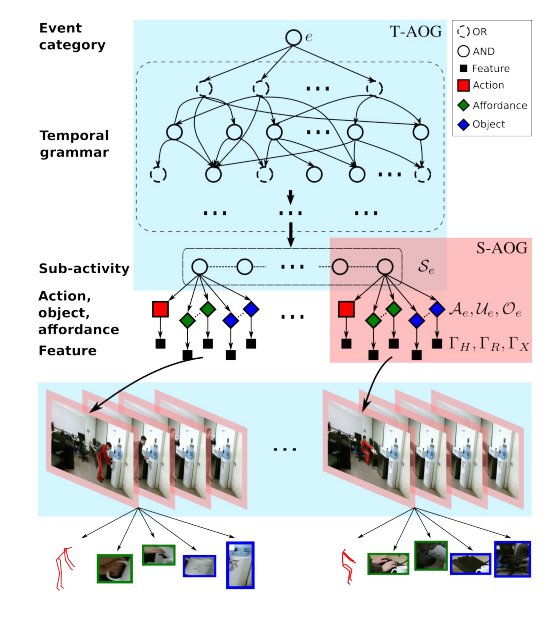}\caption{\label{fig:st-aog}ST-AOG model has two main parts: T-AOG on top represents the activity as the root, S-AOG represents the subactivities  nodes which encode actions, object affordances and interactions as the state context.\cite{Human_Activities_Grammar_2017}}.
\end{figure}

\begin{table}
\scalebox{0.82}{
\begin{tabular*}{1.2\linewidth}{@{\extracolsep{\fill}}>{\raggedright}m{.12\textwidth}
>{\raggedright}m{0.1cm}
>{\raggedright}m{0.1cm}
>{\raggedright}m{0.1cm}
>{\raggedright}m{0.1cm}
>{\raggedright}m{0.1cm}
>{\raggedright}m{0.1cm}
>{\raggedright}m{0.1cm}
>{\raggedright}m{0.1cm}
>{\raggedright}m{0.1cm}
>{\raggedright}m{0.1cm}
}

 \cline{1-11} 
 \rule{0pt}{3ex} 
\multirow{2}{*}{} & \multicolumn{2}{|l|}{Input } &  \multicolumn{2}{l|}{Features}   & \multicolumn{2}{l|}{Evaluation} & \multicolumn{2}{l|}{Training} & \multicolumn{2}{l}{Model}\tabularnewline

 \cline{2-11} 
 \rule{0pt}{3ex} 
 &  \multicolumn{1}{|c}{\rotatebox[origin=c]{90}{2D}}  &    \multicolumn{1}{c|}{\rotatebox[origin=c]{90}{3D}}  &   \multicolumn{1}{c}{\rotatebox[origin=c]{90}{   Feature learning   }} &    \multicolumn{1}{c|}{\rotatebox[origin=c]{90}{Handcrafted}} &   \multicolumn{1}{c}{\rotatebox[origin=c]{90}{Real Robot}}  &    \multicolumn{1}{c|}{\rotatebox[origin=c]{90}{Benchmark}} &   \multicolumn{1}{c}{\rotatebox[origin=c]{90}{Supervised}}  &       \multicolumn{1}{c|}{\rotatebox[origin=c]{90}{Weakly supervised }} &   \multicolumn{1}{c}{\rotatebox[origin=c]{90}{Mathematical}} &     \multicolumn{1}{l}{\rotatebox[origin=c]{90}{Neural}}\tabularnewline

 \cline{1-11} 
\rule{0pt}{3ex} 
 Koppula \etal \cite{Human_Activities_Affordance_2013}  & \multicolumn{1}{|c}{} &   \multicolumn{1}{c|}{\checkmark}  &  & \multicolumn{1}{c|}{\checkmark} & \checkmark   & \multicolumn{1}{c|}{\checkmark} &\checkmark  &  \multicolumn{1}{c|}{}  &\checkmark & \multicolumn{1}{l}{}\tabularnewline

Koppula \etal \cite{Anticipating_human_activities_affordances_2016}& \multicolumn{1}{|c}{}  &   \multicolumn{1}{c|}{\checkmark} &  &  \multicolumn{1}{c|}{\checkmark} &    &  \multicolumn{1}{c|}{\checkmark}&\checkmark  &  \multicolumn{1}{c|}{}  & \checkmark &\multicolumn{1}{l}{} \tabularnewline

 Jain \etal \cite{Structura_RNN_2016}	 &\multicolumn{1}{|c}{} & \multicolumn{1}{c|}{\checkmark}   & \checkmark &   \multicolumn{1}{c|}{}&   & \multicolumn{1}{c|}{\checkmark} &\checkmark  &  \multicolumn{1}{c|}{}  &  & \multicolumn{1}{l}{\checkmark}\tabularnewline

 Qi \etal \cite{Human_Activities_Grammar_2017} &\multicolumn{1}{|c}{} & \multicolumn{1}{c|}{\checkmark}   & \checkmark &   \multicolumn{1}{c|}{}&   & \multicolumn{1}{c|}{\checkmark} &\checkmark  &  \multicolumn{1}{c|}{}  &  & \multicolumn{1}{l}{\checkmark}\tabularnewline

 Vu \etal \cite{Predicting_actions_static_scenes_2014} &\multicolumn{1}{|c}{\checkmark } &  \multicolumn{1}{c|}{}   &  &   \multicolumn{1}{c|}{\checkmark}&   & \multicolumn{1}{c|}{\checkmark} & \checkmark &  \multicolumn{1}{c|}{}  &\checkmark  & \multicolumn{1}{l}{}\tabularnewline

 Dutta and Zielinska \cite{Action_prediction_2017} &\multicolumn{1}{|c}{ } & \multicolumn{1}{c|}{ \checkmark}   &  &   \multicolumn{1}{c|}{\checkmark}&   & \multicolumn{1}{c|}{\checkmark} &\checkmark  &  \multicolumn{1}{c|}{}  & \checkmark & \multicolumn{1}{l}{}\tabularnewline

 Shu \etal \cite{HRI_SocialAffordances_2016} &\multicolumn{1}{|c}{ } & \multicolumn{1}{c|}{ \checkmark}   &  &   \multicolumn{1}{c|}{\checkmark}&   & \multicolumn{1}{c|}{\checkmark} &\checkmark  &  \multicolumn{1}{c|}{}  & \checkmark & \multicolumn{1}{l}{}\tabularnewline

 Shu \etal \cite{social_affordance_grammar_2017} &\multicolumn{1}{|c}{ } & \multicolumn{1}{c|}{ \checkmark}   &  &   \multicolumn{1}{c|}{\checkmark}&   & \multicolumn{1}{c|}{\checkmark} &  &  \multicolumn{1}{c|}{\checkmark}  & \checkmark & \multicolumn{1}{l}{}\tabularnewline

\bottomrule
\end{tabular*}}
\caption{Comparison between affordance-based activity recognition methods}
\label{table:affordancebasedactivityrecognitioncomparison}
\end{table}

\subsection{High-level Affordance Reasoning}
\begin{figure*}
\centering \includegraphics[width=\textwidth]{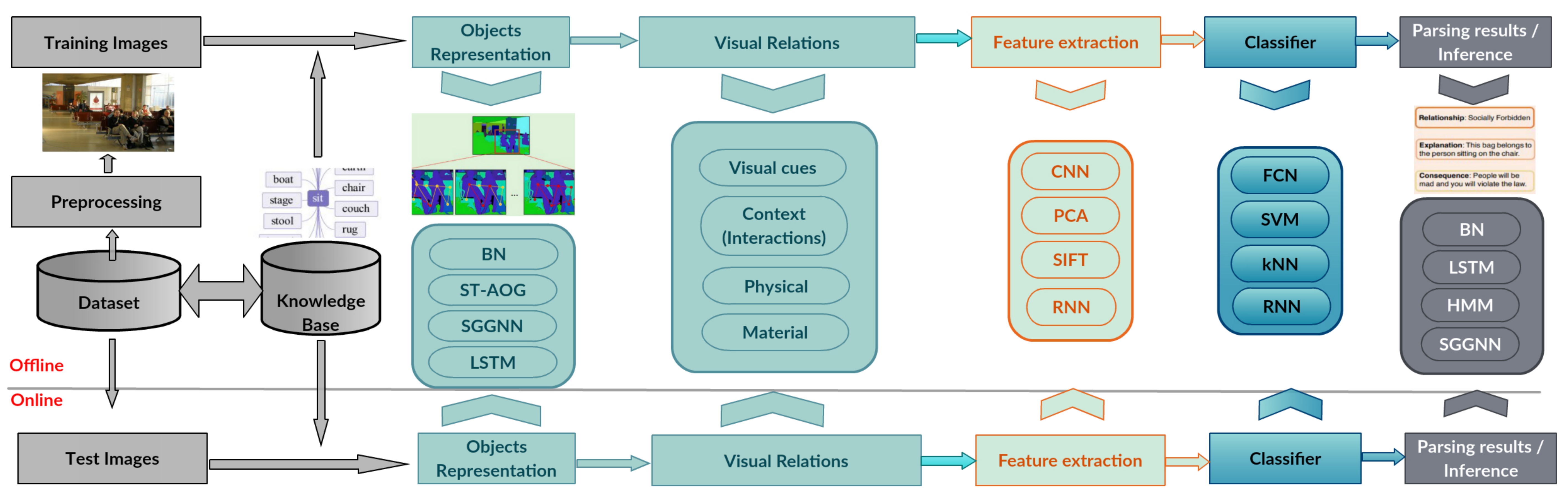}\caption{\label{fig:affordancereasoning} Affordance reasoning process. }
\end{figure*}

Affordances can be used as a tool to perform reasoning about more complex object properties and events in a scene. As an example, affordances have been used to infer the hidden properties  e.g., What is inside a container? What are the intricate relationships between objects? Or to answer complex questions about a scene. In this section, we cover research works which perform high-level reasoning based on affordances (see Figure \ref{fig:affordancereasoning}.

Zhu et al. \cite{Reasoning_Affordances_2014} proposed the first study
to discuss the visual reasoning of affordances. They developed Knowledge
base (KB) that represents the object along with other nodes which describe
attributes (Visual, Physical and Categorical) or affordances to infer
the affordance labels, human poses or relative locations. They learned
their model using Markov Logic Network (MLN) \cite{Markove_Logic_Networks_2006}
whereas zero-shot learning \cite{zero_shot_learning_2011} has been
used to predict affordances for novel objects. However, their approach
assumed that the affordance semantics and attributes are given in advance
and use static models. Chao et al. proposed this study \cite{KnowBaseAffordCat_CVPR_2015}.
They modeled the affordance semantics problems in the form of action-object
pairs as connected verb-noun nodes in WordNet \cite{WordNet_1995}
or encoding the plausibility of a matrix such as this study \cite{MatrixUsed_By_affordance_sematics_2012}.
They used novel statistical methods like co-occurrences to
infer about the affordances and give the best description (verb) for
this object(noun). Regrading reasoning about the containers' contents, G{ü}ler
et al. introduced the visual-tactile method to infer what is inside the
container. A kinect camera was used to capture the object and then
deformal model was detected before using tactile signals for reasoning.
They used a three-fingered Schunk Dextrous Hand to grasp and squeeze
the container to check whether it is full or empty. PCA was the extractor
and kNN and SVM were the classifiers of the model. However, the authors
assumed the presence of the 3D object model to perform the learning.

Yu et al. \cite{Fill_Transfer_2015} proposed a physics-based method
to reason liquid contanability affordances through two steps:
1) the best filling direction 2) considering the direction, transfer
the liquid outside to estimate the containability of an object. They
used fluid dynamics in 3D space to simulate human motion of liquid transfer to test their approach. Although this study tried to predict containability based on the visual images, it did not use the
human factors in the scene which makes the reasoning easier. Further,
they did not relate the material attributes to the reasoning e.g.
any curved tool can contain liquid even if it is made of this paper.
Seeking to overcome the human factors in the scene problem, Wang et
al. \cite{Transferring_objects_2017} developed their approach as
an enhancement of this paper \cite{Fill_Transfer_2015}. Given a 3D
scene and considering the compatability among container, containee
and pose, they learn their model to reason the best pose and container
to do the task of transferring the liquid. They provided a RGB-D dataset and they used SVM to train the approach. The process of filling
the containers is the same process of this study \cite{Fill_Transfer_2015};
they voxelize the object and simulate the filling inside its space.
In the same context, Mottaghi et al. \cite{See_Glass_Half_Full_2017}
developed an approach to reason about the affordance of liquids inside
the container (the volume, amount of liquid) and predict its
behavior. They introduced Containers Of liQuid contEnt (CODE) dataset
of RGB images along with 3D CAD models. They used deep learning in
the form of CNN and RNN to learn the containability affordance based
on contextual cues. This method depended on visual features. Phillips
et al. \cite{Seeing_glassware_2016} introduced a method to detect,
localize and segment pose estimation of transparent objects
like glasses; and they provided an annotated dataset of transparent
objects. Liang et al. studied the human cognition of the containers
to infer its affordances (object's containment and number of objects
that can be contained inside) \cite{Containing_Relations_Physical_Simulation_2015}.
In a recent study, Liang et al. \cite{What_is_Where_2016} inferred containment affordances and relations in RGB-D videos
over time. For example, the fridge contains the eggs carton which contains
the eggs. They used the human actions (move-in, move-out, no-change
and paranormal-change) to draw containment graphs based on spatial-temporal
relations; that is, the action used to detect containment objects.
They introduced RGB-D videos dataset and they developed probabilistic
dynamic programming to optimize the containment graphs. Zhu et al. \cite{Inferring_forces_videos_2016} used physics-based simulation to infer the forces and pressures
for the different body parts while sitting on a chair. They predicted
the object affordances through human utilities while sitting e.g.
comfortable and lazy. Krunic et al. introduced a model to include
verbal information to link between utterances and objects through
inferring the context between words, actions and its outcomes. Zhu
et al. \cite{VAQueries_KB_2015} built a knowledge base (KB) to reason
answers for image questions. They represented nodes in their model
as attributes, affordance labels, scene categories or image features.
In their most recent study, Chuang et al. presented a promising study to reason
about the action-object affordances based on physical and social norms
\cite{Learning_ActProperly_2017}. They annotated the ADE20k \cite{ADE20K_Dataset_Object_Parts_2016}
dataset with affordance features, detailed explanation and potential
consequences for every object. For instance, pouring water into a
cup has explanation that it is improper to pour because the cup is
full and consequence that you will make a mess in that place. They built
the model using a Gated Graph Neural Network (GGNN) while Spatial
version of GGNN has been employed to reason affordances. This study
combined the social norms, physical features, visual attributes and
situation parameters to infer the affordances and its relations whether
positive or negative. To summarize, Table \ref{table:affordancereasoningcomparison} gives more details about the used methods in the literature of reasoning.

\begin{figure}
\centering \includegraphics[width=0.95\columnwidth,height=0.12\paperheight]{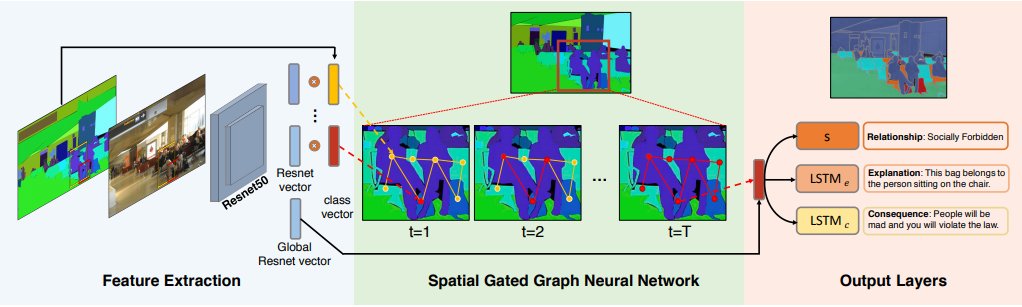}\caption{\label{fig:learntoact} The deep learning architecture that is proposed by \cite{Learning_ActProperly_2017}. ResNet is used as the feature extraction layer and the results are fed into a customized Spatial Gated Graph Neural Network (SGGNN) to represent the visual affordances through graphs. Hence, use this graph to reason the affordances, explanations and consequences}.
\end{figure}

\begin{table}
\scalebox{0.82}{
\begin{tabular*}{1.2\linewidth}{@{\extracolsep{\fill}}>{\raggedright}m{.13\textwidth}
>{\raggedright}m{0.1cm}
>{\raggedright}m{0.1cm}
>{\raggedright}m{0.1cm}
>{\raggedright}m{0.1cm}
>{\raggedright}m{0.1cm}
>{\raggedright}m{0.1cm}
>{\raggedright}m{0.1cm}
>{\raggedright}m{0.1cm}
>{\raggedright}m{0.1cm}
>{\raggedright}m{0.1cm}
>{\raggedright}m{0.1cm}
>{\raggedright}m{0.1cm}
}
\toprule 
\multirow{2}{*}{Methods} & \multicolumn{3}{|c|}{Input } &  \multicolumn{2}{c|}{Features}   & \multicolumn{2}{c|}{Evaluation} & \multicolumn{2}{c|}{Training} & \multicolumn{2}{c}{Model}\tabularnewline
  
 \cline{2-12} 
 &  \multicolumn{1}{|c}{\rotatebox[origin=c]{90}{2D}}  &   \rotatebox[origin=c]{90}{3D} &    \multicolumn{1}{c|}{\rotatebox[origin=c]{90}{Multimodal}}   &  \rotatebox[origin=c]{90}{  Feature learning   } &   \multicolumn{1}{c|}{\rotatebox[origin=c]{90}{Handcrafted}} & \rotatebox[origin=c]{90}{Simulation}&   \multicolumn{1}{c|}{\rotatebox[origin=c]{90}{Benchmark}} &   \rotatebox[origin=c]{90}{Supervised}   &    \multicolumn{1}{c|}{\rotatebox[origin=c]{90}{Unsupervised}} &   \rotatebox[origin=c]{90}{Mathematical} &     \multicolumn{1}{l}{\rotatebox[origin=c]{90}{Neural}}\tabularnewline
 \midrule 

 Zhu \etal \cite{Reasoning_Affordances_2014} & \multicolumn{1}{|c}{\checkmark} &  & \multicolumn{1}{c|}{}  & & \multicolumn{1}{c|}{\checkmark }   & & \multicolumn{1}{c|}{\checkmark } & \checkmark  &\multicolumn{1}{c|}{}  &\checkmark  & \multicolumn{1}{l}{}\tabularnewline
  
  Chao \etal \cite{KnowBaseAffordCat_CVPR_2015} & \multicolumn{1}{|c}{\checkmark} &  & \multicolumn{1}{c|}{}  & & \multicolumn{1}{c|}{\checkmark }    & & \multicolumn{1}{c|}{\checkmark } & \checkmark  &\multicolumn{1}{c|}{}  &\checkmark  & \multicolumn{1}{l}{}\tabularnewline
   
    Yu \etal \cite{Fill_Transfer_2015} & \multicolumn{1}{|c}{} & \checkmark & \multicolumn{1}{c|}{}  & & \multicolumn{1}{c|}{\checkmark }   & & \multicolumn{1}{c|}{\checkmark } & \checkmark  &\multicolumn{1}{c|}{}  &\checkmark  & \multicolumn{1}{l}{}\tabularnewline

          Zhu \etal \cite{Inferring_forces_videos_2016}& \multicolumn{1}{|c}{} & \checkmark & \multicolumn{1}{c|}{}  & & \multicolumn{1}{c|}{\checkmark } &    \checkmark & \multicolumn{1}{c|}{ } & &\multicolumn{1}{c|}{ \checkmark }  &\checkmark  & \multicolumn{1}{l}{}\tabularnewline
       
    Wang \etal \cite{Transferring_objects_2017}  & \multicolumn{1}{|c}{} & \checkmark & \multicolumn{1}{c|}{}  & & \multicolumn{1}{c|}{\checkmark } &   & \multicolumn{1}{c|}{\checkmark } & \checkmark  &\multicolumn{1}{c|}{}  &\checkmark  & \multicolumn{1}{l}{}\tabularnewline
        
       Liang \etal \cite{What_is_Where_2016}  & \multicolumn{1}{|c}{} & \checkmark & \multicolumn{1}{c|}{}  & & \multicolumn{1}{c|}{\checkmark }    & & \multicolumn{1}{c|}{\checkmark } & \checkmark  &\multicolumn{1}{c|}{}  &\checkmark  & \multicolumn{1}{l}{}\tabularnewline
       
 Mottaghi \etal \cite{See_Glass_Half_Full_2017} & \multicolumn{1}{|c}{\checkmark}  & &   \multicolumn{1}{c|}{} & \checkmark  &  \multicolumn{1}{c|}{} & &  \multicolumn{1}{c|}{\checkmark}& \checkmark  &\multicolumn{1}{c|}{}  &  &\multicolumn{1}{l}{\checkmark} \tabularnewline
 
 Phillips \etal \cite{Seeing_glassware_2016} &\multicolumn{1}{|c}{\checkmark } & &\multicolumn{1}{c|}{ }   &  &   \multicolumn{1}{c|}{\checkmark}&   & \multicolumn{1}{c|}{\checkmark} & &\multicolumn{1}{c|}{\checkmark}  & \checkmark & \multicolumn{1}{l}{}\tabularnewline
 
 Zhu \etal \cite{VAQueries_KB_2015} & \multicolumn{1}{|c}{}  & &   \multicolumn{1}{c|}{\checkmark} & \checkmark  &  \multicolumn{1}{c|}{}    & &  \multicolumn{1}{c|}{\checkmark}& \checkmark  &\multicolumn{1}{c|}{}  &  &\multicolumn{1}{l}{\checkmark} \tabularnewline

  Chuang \etal \cite {Learning_ActProperly_2017}& \multicolumn{1}{|c}{\checkmark}  & &   \multicolumn{1}{c|}{} & \checkmark  &  \multicolumn{1}{c|}{}  & &  \multicolumn{1}{c|}{\checkmark}& \checkmark  &\multicolumn{1}{c|}{}  &  &\multicolumn{1}{l}{\checkmark} \tabularnewline

\bottomrule
\end{tabular*}}
\caption{Comparison between affordance reasoning methods}
\label{table:affordancereasoningcomparison}
\end{table}

\section{Functional Scene Understanding}\label{Functional SU}

\begin{figure}[ht]
\centering \includegraphics[width=.8\columnwidth,height=0.23\paperheight]{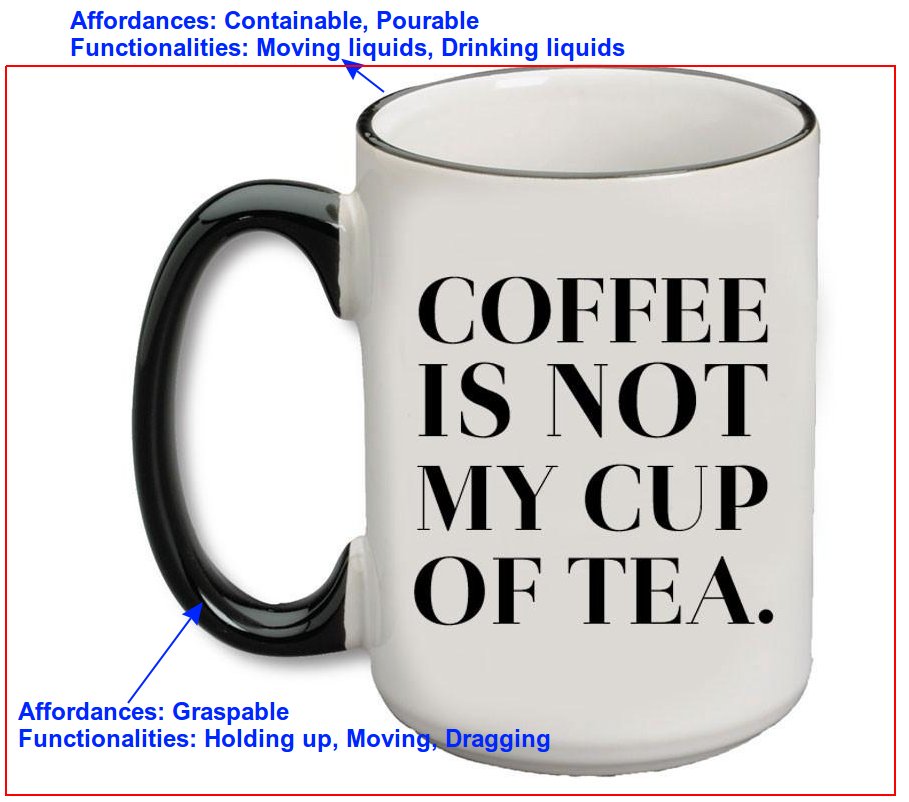}\caption{\label{fig:multifuncti} The relation between functions and affordances is many to many.}
\end{figure}

Zhu \etal \cite{Understanding_tools_Affordances_2015} made a distinction between functionality and affordances. The problem of affordance particularly depends on detecting and classifying objects; learning their affordances; and performing more detailed understanding and reasoning based on the learned affordance properties. The problem of understanding the functions is related to affordance learning, but specifically aims to identify the tasks that can be performed with an object. In contrast, affordance learning reasons about object functions in the context of agent (animal or robot). Further more, some objects have both affordance parts and function parts e.g. the hammer has a head which is suitable for striking objects (function) and a handle from which it can be grasped (affordance). Some of these object parts have multiple affordances but a single function e.g. the \emph{hammer handle} can be used push or pull objects so it has pushable and pullable affordances  while the function is head support. In contrast, the \emph{hammer claws}  have multiple functions and a multiple affordances, i.e., claws can function as a lever and also can be used to pull out nails from timber but has the affordance of hooking, grasping and pushing. Thus, the relation between the affordance and the function is many to many as Figure \ref{fig:multifuncti} shows. 

\begin{figure*}
\centering \includegraphics[width=\linewidth,height=0.1\paperheight]{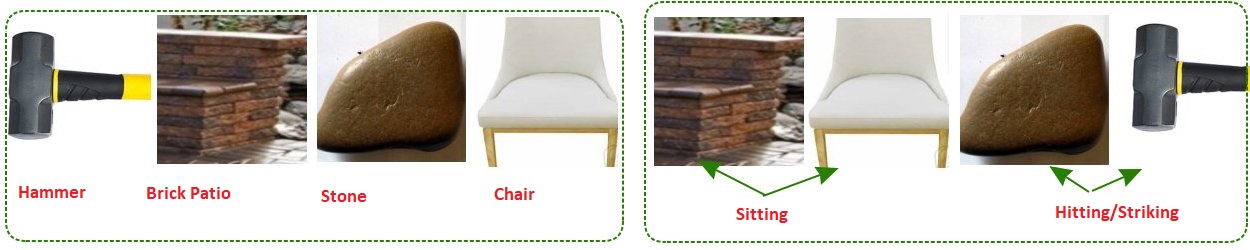}
\caption{Categories according to the functionality. The left side shows the traditional categorizing or object-labels categorization whereas the right side categorizes the objects in functional manner. For example, handle-graspable category includes the first four objects from the right side}\label{fig:Category_Functionality}
\end{figure*}

As scene understanding is an old problem, it has been studied from many perspectives. However, functional scene understanding has not been thoroughly investigated before. Although this problem is of high significance for robotics, a few research efforts have focused on functional scene understanding. For example, a robot can not clean the kitchen dynamically without understanding how to use taps or electricity plugs to run the vacuum. So that the functional scene understanding
is crucial to the robots particularly cognitive robots. As Figure \ref{fig:Category_Functionality} shows, categorizing the images according to its purpose is meaningful in many cases than categorizing according to its appearance.

\subsection{Function-parts recognition}

\begin{figure}
\centering \includegraphics[width=\linewidth,height=0.35\paperheight]{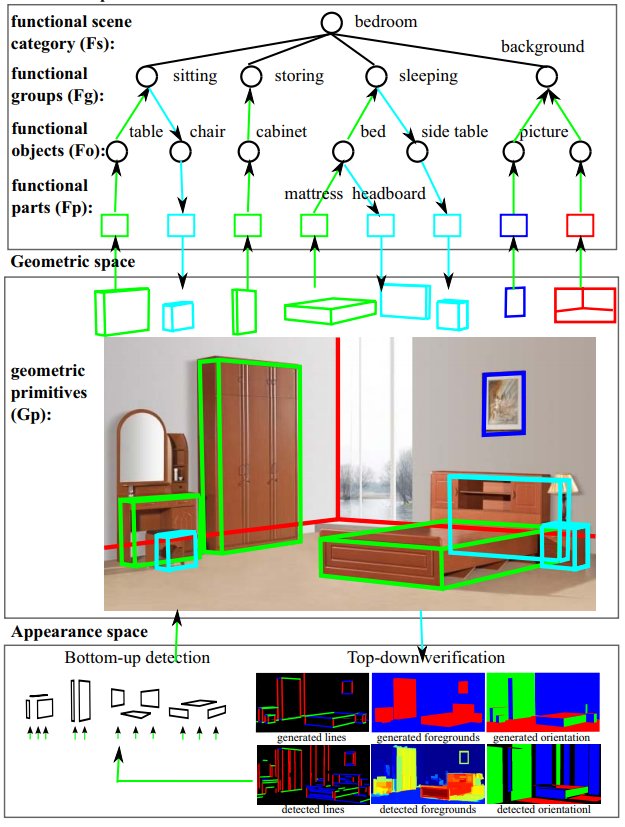}
\caption{Function-Geometry Appearance (FGA) use functional cues (categories, objects and parts), geometrical features (e.g. object 3D size and corners values ) and appearance parameters to enhance the prediction accuracy \cite{scene_Parsing_function_geometry_appearance_2013}.}\label{fig:functionaldescriptor}
\end{figure}

Inspired by Gibson \cite{GibsonAffordanceWord_1977}, the authors in \cite{Physical_from_function, Association_Func_structure_1991} looked for the relation between recognizing objects and its functions. In other words, they aim to recognize the objects according to their usage features instead of their visual properties. Rivlin \etal used the relations of objects (e.g. size and orientation) and parts to reason about the functionality \cite{Recognition_By_Functional_Parts_1995}. Desai and Ramanan performed an interesting study to predict functional regions and functional landmarks in an image \cite{Functional_regions_2013}. They used deformable part models (DPM) \cite{Detection_DPM_2010} and\cite{Phraselets_2012} as pose detector to extract spatial relations of objects followed  by kNN to predict functional regions in the scene. For instance, the functional part of the vacuum cleaner is the power buttons. They targeted to detect 3D objects based on their functional parts, spatial relations and functional landmarks and tested their approach on the attributes of people dataset  \cite{Detectio_Attr_action_poselets_2013}. Zhao and Zhu proposed a stochastic method, Function-Geometry Appearance (FGA), to parse 3D scenes by combining features of the functionality, geometry and appearance \cite{scene_Parsing_function_geometry_appearance_2013}.
They modeled the FGA through top-down/bottom-up hierarchy and used MCMC to build the algorithm and infer the parse tree (see Figure \ref{fig:functionaldescriptor}). The functional descriptor was composed of functional scene categories, groups, objects and parts and use AND-OR rules to understand the affordance and therefore the full scene.

Shiraki \etal \cite{Grasping_FunctionalUnites_2014} introduced the differentiation between main parts and subordinate parts (like hammer head and hammer handle) which is important to reason about object functions.  Zhu \etal \cite{Understanding_tools_Affordances_2015}
used a simulation engine to perform affordance and functionality analysis. They differentiated between functional learning which was defined as the right location to do a task on the target object; and affordance learning which was defined as the best location to grasp the object depending on the tool type. Additionally, they fused physical features (e.g. forces, pressure
and volume), human pose from imagined action, affordance features and functional features together to understand and infer tool's usage. Ye \etal\cite{DFSU2017} addressed object function detection by introducing a novel approach through three main steps: selective search to detect object proposals; feature extraction by VGG-F and VGG-S network and training the neural network with stochastic gradient descent to perform functional object detection. The authors inspired that problem, but it has the following limitations: (a) the accuracy has not exceeded 25\%, (b) it requires long time to train because of using three different stages i.e. detecting, feature extraction and training. Since this problem is relatively new, it needs many further investigations to achieve a practical and successful baseline. 
In a different way of processing, Pechuk \etal structured their problem into function classification model in 3D images, where the category labels were defined to be e.g., ``to sit" and ``to use" \cite{Function_Based_Classification_2008}. Their scheme operates on a  multi-level hierarchy of parts that have various functionality signatures. They named these parts as primitives and classified each part or group of them as functional part if they offer the same functionality. To link between primitive and functional parts, they defined association, connection and mapping relationships. Kitano et al. \cite{Estimate_Functions_CNN_2016} build a CNN to estimate the object functions based on their appearance.
As \cite{WhatMakesChairAChair_2011}, the usage of CAD models is repeated in \cite{Predicting_object_functionality_2013} while Hinkle and Olson proposed learning method for robots that face novel (unknown) objects. Based on physical simulations of a falling sphere onto objects, the robot can classify the objects according to functionality or potential capabilities. They learned three functions: drinking vessel, table and sittable. Awaad et al. \cite{functional_affordances_socializing_robots_2015} proposed using functional affordances to serve social robots. They challenge that using functional affordances of objects represent the start point for socializing robots to achieve the goal. For example, using a mug instead of the glass, which is not available, can achieve the goal of drinking water. To learn functional affordances, they built an ontology for the objects usage and ranked them according to whether it is the primary function or a secondary function.  Madry et al. proposed categorizing the 2D-3D images according to its functionality to reason about grasp planning \cite{object_categories_reasoning_2012}. They modeled their approach as a Bayesian Network and extended this probabilistic method \cite{Task_constraint_grasping_2010} for reasoning. They used five functional categories to perform tasks as follows: hand-over, pouring, dish washing, playing and tool-use. Xie et al. \cite{Inferring_Dark_matters_2013} named
functional objects as \textquotedbl{}dark matter\textquotedbl{} and all the functional objects in a scene as \textquotedbl{}dark energy\textquotedbl{}; and proposed a Bayesian framework to represent their model whereas Markov Chain Monte Carlo (MCMC) is used for inference. In contrast to the previous
works, this method used the human trajectories in order to localize the functional objects from videos. 
In the context of functional categorization, Gall \etal proposed a novel method to solve the problem of unsupervised categorization of objects based on functionality. They inferred the functionality from motion of human-object interactions. Furthermore, they extracted the human poses through interaction, encoded a string of poses and measured the similarity between the actions with Levenshtein distance method \cite{Levenshtein_Similarity_measure_1996}. Saponaro \etal \cite{human_intentions_gesture_2013} presented a geometric transformation method to the hand actions (push, grasp and tap) to predict physical actions and reason intentions of the human. They built their model based on the affordance relationships mentioned in Figure \ref{fig:objects_actions_Image}.

\subsection{Function from Motion}
Gupta \etal \cite{spatial_functional_compatibility_2009} used spatial and functional features to understand human actions in static images and videos. They argued that the functionality, which has been pointed
out as motion trajectories and human poses, and pose can interpret the scene in terms of action recognition. They built Bayesian Network model which fused functional and spatial data for object, action and object effect for recognition.
 Oh \etal \cite{Functional_objects_scene_context_2010} detected the functionality of moving objects in order to understand the scene context. They claimed that appearance features are not enough for some objects in traffic scenes (e.g. parks and way lanes) because all of them are similar without much difference. Turek \etal \cite{unsupervised_functional_categories_2010} presented a novel method to categorize a scene according to the motion of objects. The categories are function-based such as sidewalks or parking areas. Local descriptors to model object properties, behavioral relationships and temporal features were used to discriminate between functional regions. Inspired by \cite{unsupervised_functional_categories_2010}, Zen \etal used the same steps but additionally included the semantics of the moving object to get a more  informative descriptor \cite{semantic_functional_descriptors_categorization_2012}. Additionally, they tested their model  on a traffic dataset \cite{Traffic_dataset_2011}.  Rhinehart and Kitani \cite{First_Person_vision_2016} analyzed egocentric videos to extract functional descriptors to learn \textquotedbl{}Action
Maps\textquotedbl{} (six actions) \cite{Action_Maps_2014} which can tell a user about how to perform activities in novel scenes.  Pirk \etal \cite{Understanding_exploiting_object_interaction_landscapes_2017} used scanners to track interaction parts and built the functionality descriptor by analyzing motion of these parts interaction with an object e.g. analyzing an agent tries to pour water into a mug. 


\subsection{Function from Interactions}
 Gupta \etal \cite{Human_workspace_affordance_2011} modeled human-scene interaction with a functional descriptor in terms of 3D human poses to predict various poses of the human and they named it as the human workspace. They used a 3D cuboidal model to fit objects and predicted the subjective affordances. In \cite{Scene_Semantics_2012}, Delaitre \etal functional descriptors were proposed to allow a better scene understanding. The main idea is to extract functional descriptors from human-object interaction aiming to achieve better object recognition results. They extended an existing dataset \cite{People_Watching_2012_Dataset} to evaluate their proposed model. Similarly, Fouhey \etal used the idea of \cite{spatial_functional_compatibility_2009} to predict the affordance labels from videos by observing person-object interactions \cite{People_Watching_2012_Dataset}. For instance, the knife has the \textquotedbl{}cut\textquotedbl{}  function because many persons used it for cutting. Additionally, they provided a video dataset for indoor scenes with function labels.  In \cite{spatio_temporal_object_object_relationships_2014}, Pieropan \etal proposed using object-to-object spatio-temporal relationships to create a so called ``object context" along with functional descriptor to predict the human activities. As an example, only the presence of a mug does not confirm if a drinking action could take place, but the presence of a water bottle beside it increases the likelihood of a drinking action. They used the kitchen images of the dataset proposed in \cite{Function_Object_Activity_decriptor_2013} and trained a probabilistic model, Conditional Random Field (CRF), to classify functional classes into four types: tools, ingredients, support and containers. Yao \etal represented function understanding problem as weakly supervised to discover all possible functionalities for each object \cite{DiscoverObjectFunction_2013}. They used unsupervised clustering in an iterative manner to categorize human-object interactions, then used these updates as input for detection and pose estimation, and finally discovered the functionalities. The authors tested their model through musical instrument dataset \cite{Dataset_human_object_interaction_2010} which contains images for human-object interactions. 
Given human-object interaction observations, Stark  \etal \cite{Grasping_functionalObectsonlearnedcues_2008} proposed a learning mechanism to categorize grasping affordances as object functionalities as shown in Figure \ref{fig:Category_Functionality}. Through observing the interaction with some object, the affordance cues are defined. In other words, the affordance has been defined as relation between robot hand and an object. The implicit shape model (ISM) \cite{ISM_categorization_segmentation_2006} was the main algorithm used by \cite{Grasping_functionalObectsonlearnedcues_2008} to categorize the objects. To this end, these cues have been used to predict the grasping points of a 2D image. Pieropan et al. \cite{Function_Object_Activity_decriptor_2013}, applied functional descriptors/cues, which have been learned from hand-object interaction, to understand human activities from RGB-D videos. They represented the objects by their interaction with human hands as well as they encoded these objects as strings through which the string kernel \cite{String_Kernels_2002} measured their similarity. Likewise, they used spatial location and temporal trajectory to estimate object position relative to the hand. Hence, the estimated object position produced a so called functional descriptor. Thereafter, they fused the similarity measures with functional descriptors to recognize human activities. Mar et al. \cite{Affordance_Learning_multimodal_2015} proposed a method to learn tool affordances based on its function and the way of interaction (grasping). To find functional descriptors, they learned geometrical features through Self-Organized Maps (SOM) and K-means.
Hu et al. introduced what called Interaction CONtext (ICON) to describe the functionality of 3D object through geometric features \cite{ICON_2015} focused on other aspects of functionality usage in vision research. The main idea was to define what is called the contextual descriptor to describe the shape functionality in the presence of other objects using Interaction Bisector Surface (IBS) \cite{Interaction_Bisector_Surface_2014}. In other words, they used object-to-object interaction to built a geometrical descriptor for  functionality analysis and hence they recognized the correspondences between similar parts
on various shapes of 3D images. They used the Trimble 3D Warehouse\footnote{https://3dwarehouse.sketchup.com/} to test their experiment. Savva et al. \cite{Understanding_exploiting_object_interaction_landscapes_2017} built action maps for potential actions through scanning
geometry of the captured 3D scenes, reconstructing depth meshs, tracking
human interactions to define the functionality descriptors and therefore predicting the affordances of unseen objects. 

\subsection{Functional descriptor shape and correspondence}
Jain \etal \cite{BN_tool_representation_2013} proposed a probabilistic model based on geometric features, which is related to the functionalities such as those based on material and shape, along with probabilistic dependencies between the effects, tool's actions and tool's functional
features. A Bayesian Network (BN) is used in their scheme because of  its ability to handle the probabilistic dependencies between nodes.
Laga \etal proposed a model that extracted the pair-wise semantics of shapes through combining structural and geometric features \cite{semantic_Correspondence_Functionality_2013}. Additionally, they recognized the functionality of the shapes (e.g. graspable and container) using supervised learning methods. Kim \etal used affordances as priori information to predict the correspondence in human poses through which they predicted the functional descriptors \cite{Shape2Pose_2014}. Apart from above-mentioned studies which used abstract functionality
such as \textquotedbl{}to pour\textquotedbl{} and \textquotedbl{}to move\textquotedbl{}. 
In recent study, Lun \etal \cite{Functionality_preserving_2017} designed a unified model to detect a human pose according to human-object affordances (leaing, holding, sitting and treating) along with object parts. The functionality descriptor has been employed to recover mechanical assemblies or parts from raw scans \cite{Recover_Functional_Parts_2018}. They used segmentation and joint optimization to learn their scheme. Hu \etal in recent study \cite{Co-analysis_Interactions_2016} proposed a method to analyze inter-object relations and intra-object relation aiming to categorize the objects based on their functionalities. They used objects' parts contexts, semantics and functionalities to recognize their shapes.\\ Various methods with different ways of representation have been introduced to address the functionality issues. Table \ref{table:functionscenecomparison} summarizes and compares the most important properties to make it more understandable. 

\begin{table}
\scalebox{0.8}{
\begin{tabular*}{1.25\linewidth}{@{\extracolsep{\fill}}>{\raggedright}m{.28\linewidth}
>{\raggedright}m{0.01\linewidth}
>{\raggedright}m{0.01\linewidth}
>{\raggedright}m{0.01\linewidth}
>{\raggedright}m{0.01\linewidth}
>{\raggedright}m{0.01\linewidth}
>{\raggedright}m{0.01\linewidth}
>{\raggedright}m{0.01\linewidth}
>{\raggedright}m{0.01\linewidth}
>{\raggedright}m{0.01\linewidth}
>{\raggedright}m{0.01\linewidth}
>{\raggedright}m{0.01\linewidth}
>{\raggedright}m{0.01\linewidth}
}
\toprule 
\multirow{2}{*}{} & \multicolumn{2}{|l|}{Input } &  \multicolumn{2}{l|}{Features}   & \multicolumn{3}{l|}{Evaluation} & \multicolumn{3}{l|}{Training} & \multicolumn{2}{l}{Abstraction}\tabularnewline
  
 \cline{2-13} 
 &  \multicolumn{1}{|c}{\rotatebox[origin=c]{90}{2D}}  &       \multicolumn{1}{c|}{\rotatebox[origin=c]{90}{3D}}   &  \rotatebox[origin=c]{90}{  Feature learning   } &   \multicolumn{1}{c|}{\rotatebox[origin=c]{90}{Handcrafted}} &   \rotatebox[origin=c]{90}{Real Robot}  & \rotatebox[origin=c]{90}{Simulation}&   \multicolumn{1}{c|}{\rotatebox[origin=c]{90}{Benchmark}} &   \rotatebox[origin=c]{90}{Supervised}   &\rotatebox[origin=c]{90}{  Weakly supervised  } &    \multicolumn{1}{c|}{\rotatebox[origin=c]{90}{Unsupervised}} &   \rotatebox[origin=c]{90}{Mathematical} &     \multicolumn{1}{l}{\rotatebox[origin=c]{90}{Neural}}\tabularnewline
 \midrule 

 Zhu \etal \cite{Understanding_tools_Affordances_2015} & \multicolumn{1}{|c}{}  & \multicolumn{1}{c|}{\checkmark}  & & \multicolumn{1}{c|}{\checkmark } &   & & \multicolumn{1}{c|}{\checkmark } & \checkmark & &\multicolumn{1}{c|}{}  &\checkmark  & \multicolumn{1}{l}{}\tabularnewline
  
  Shiraki \etal \cite{Grasping_FunctionalUnites_2014} & \multicolumn{1}{|c}{}  & \multicolumn{1}{c|}{\checkmark}  & & \multicolumn{1}{c|}{\checkmark } &   & & \multicolumn{1}{c|}{\checkmark } & & &\multicolumn{1}{c|}{\checkmark }  &\checkmark  & \multicolumn{1}{l}{}\tabularnewline

    Turek \etal \cite{unsupervised_functional_categories_2010} & \multicolumn{1}{|c}{\checkmark}  & \multicolumn{1}{c|}{}  & & \multicolumn{1}{c|}{\checkmark } &   & & \multicolumn{1}{c|}{\checkmark } & & &\multicolumn{1}{c|}{\checkmark }  &\checkmark  & \multicolumn{1}{l}{}\tabularnewline

    Yao \etal \cite{DiscoverObjectFunction_2013} & \multicolumn{1}{|c}{\checkmark}  & \multicolumn{1}{c|}{}  & & \multicolumn{1}{c|}{\checkmark } &   & & \multicolumn{1}{c|}{\checkmark } & & \checkmark&\multicolumn{1}{c|}{ }  &\checkmark  & \multicolumn{1}{l}{}\tabularnewline
  
  Pechuk \etal \cite{Function_Based_Classification_2008} & \multicolumn{1}{|c}{}  & \multicolumn{1}{c|}{\checkmark}  & & \multicolumn{1}{c|}{\checkmark } &   & & \multicolumn{1}{c|}{\checkmark } & \checkmark & &\multicolumn{1}{c|}{}  & & \multicolumn{1}{l}{\checkmark }\tabularnewline

  Hinkle and Olson \cite{Predicting_object_functionality_2013} & \multicolumn{1}{|c}{}  & \multicolumn{1}{c|}{\checkmark}  & & \multicolumn{1}{c|}{\checkmark } &   & & \multicolumn{1}{c|}{\checkmark } & \checkmark & &\multicolumn{1}{c|}{}  &\checkmark  & \multicolumn{1}{l}{}\tabularnewline

Jain \etal\cite{BN_tool_representation_2013}  & \multicolumn{1}{|c}{\checkmark}  & \multicolumn{1}{c|}{}  & & \multicolumn{1}{c|}{\checkmark } &   &\checkmark & \multicolumn{1}{c|}{ } &  & &\multicolumn{1}{c|}{\checkmark}  &\checkmark  & \multicolumn{1}{l}{}\tabularnewline

Awaad et al. \cite{functional_affordances_socializing_robots_2015}  & \multicolumn{1}{|c}{\checkmark}  & \multicolumn{1}{c|}{}  & & \multicolumn{1}{c|}{\checkmark } &   &\checkmark & \multicolumn{1}{c|}{ } &  & &\multicolumn{1}{c|}{\checkmark}  &\checkmark  & \multicolumn{1}{l}{}\tabularnewline

 Madry et al. \cite{object_categories_reasoning_2012} & \multicolumn{1}{|c}{}  & \multicolumn{1}{c|}{\checkmark}  & & \multicolumn{1}{c|}{\checkmark } &   & & \multicolumn{1}{c|}{\checkmark } & \checkmark & &\multicolumn{1}{c|}{}  &\checkmark  & \multicolumn{1}{l}{}\tabularnewline
 
 Xie et al. \cite{Inferring_Dark_matters_2013}  & \multicolumn{1}{|c}{\checkmark}  & \multicolumn{1}{c|}{}  & & \multicolumn{1}{c|}{\checkmark } &   & & \multicolumn{1}{c|}{\checkmark } & \checkmark & &\multicolumn{1}{c|}{}  &\checkmark  & \multicolumn{1}{l}{}\tabularnewline
  
 Oh \etal \cite{Functional_objects_scene_context_2010}  & \multicolumn{1}{|c}{\checkmark}  & \multicolumn{1}{c|}{}  & & \multicolumn{1}{c|}{\checkmark } &   & & \multicolumn{1}{c|}{\checkmark } & \checkmark & &\multicolumn{1}{c|}{}  &\checkmark  & \multicolumn{1}{l}{}\tabularnewline

 Zen \cite{semantic_functional_descriptors_categorization_2012} \etal
& \multicolumn{1}{|c}{\checkmark}  & \multicolumn{1}{c|}{}  & & \multicolumn{1}{c|}{\checkmark } &   & & \multicolumn{1}{c|}{\checkmark } & \checkmark & &\multicolumn{1}{c|}{}  &\checkmark  & \multicolumn{1}{l}{}\tabularnewline

Gupta \etal \cite{spatial_functional_compatibility_2009} 
& \multicolumn{1}{|c}{\checkmark}  & \multicolumn{1}{c|}{}  & & \multicolumn{1}{c|}{\checkmark } &   & & \multicolumn{1}{c|}{\checkmark } & \checkmark & &\multicolumn{1}{c|}{}  &\checkmark  & \multicolumn{1}{l}{}\tabularnewline

 Delaitre  \etal \cite{Scene_Semantics_2012}
& \multicolumn{1}{|c}{\checkmark}  & \multicolumn{1}{c|}{}  & & \multicolumn{1}{c|}{\checkmark } &   & & \multicolumn{1}{c|}{\checkmark } & \checkmark & &\multicolumn{1}{c|}{}  &\checkmark  & \multicolumn{1}{l}{}\tabularnewline

Fouhey \etal \cite{People_Watching_2012_Dataset}
& \multicolumn{1}{|c}{\checkmark}  & \multicolumn{1}{c|}{}  & & \multicolumn{1}{c|}{\checkmark } &   & & \multicolumn{1}{c|}{\checkmark } & \checkmark & &\multicolumn{1}{c|}{}  &\checkmark  & \multicolumn{1}{l}{}\tabularnewline

Rhinehart and Kitani \cite{First_Person_vision_2016}
& \multicolumn{1}{|c}{\checkmark}  & \multicolumn{1}{c|}{}  & & \multicolumn{1}{c|}{\checkmark } &   & & \multicolumn{1}{c|}{\checkmark } & \checkmark & &\multicolumn{1}{c|}{}  &\checkmark  & \multicolumn{1}{l}{}\tabularnewline

Zhao and Zhu \cite{scene_Parsing_function_geometry_appearance_2013}
& \multicolumn{1}{|c}{\checkmark}  & \multicolumn{1}{c|}{}  & & \multicolumn{1}{c|}{\checkmark } &   & & \multicolumn{1}{c|}{\checkmark } & \checkmark & &\multicolumn{1}{c|}{}  &\checkmark  & \multicolumn{1}{l}{}\tabularnewline

 Kim \etal \cite{Shape2Pose_2014} & \multicolumn{1}{|c}{}  & \multicolumn{1}{c|}{\checkmark}  & & \multicolumn{1}{c|}{\checkmark } &   & & \multicolumn{1}{c|}{\checkmark } & \checkmark & &\multicolumn{1}{c|}{}  &\checkmark  & \multicolumn{1}{l}{}\tabularnewline
 
  Hu \etal \cite{ICON_2015} & \multicolumn{1}{|c}{}  & \multicolumn{1}{c|}{\checkmark}  & & \multicolumn{1}{c|}{\checkmark } &   & & \multicolumn{1}{c|}{\checkmark } & \checkmark & &\multicolumn{1}{c|}{}  &\checkmark  & \multicolumn{1}{l}{}\tabularnewline
  
   Laga \etal \cite{semantic_Correspondence_Functionality_2013} & \multicolumn{1}{|c}{}  & \multicolumn{1}{c|}{\checkmark}  & & \multicolumn{1}{c|}{\checkmark } &   & & \multicolumn{1}{c|}{\checkmark } & \checkmark & &\multicolumn{1}{c|}{\checkmark}  &\checkmark  & \multicolumn{1}{l}{}\tabularnewline
   
  Lun \etal \cite{Functionality_preserving_2017} & \multicolumn{1}{|c}{}  & \multicolumn{1}{c|}{\checkmark}  & & \multicolumn{1}{c|}{\checkmark } &   & \checkmark& \multicolumn{1}{c|}{ } & \checkmark & &\multicolumn{1}{c|}{}  &\checkmark  & \multicolumn{1}{l}{}\tabularnewline
  
Savva et al.  \cite{Action_Maps_functionality_2014}& \multicolumn{1}{|c}{}  & \multicolumn{1}{c|}{\checkmark}  & & \multicolumn{1}{c|}{\checkmark } &   & & \multicolumn{1}{c|}{\checkmark } & \checkmark & &\multicolumn{1}{c|}{}  &\checkmark  & \multicolumn{1}{l}{}\tabularnewline

 Lun \etal \cite{Functionality_preserving_2017} & \multicolumn{1}{|c}{}  & \multicolumn{1}{c|}{\checkmark}  & & \multicolumn{1}{c|}{\checkmark } &   & & \multicolumn{1}{c|}{\checkmark } & \checkmark & &\multicolumn{1}{c|}{}  &\checkmark  & \multicolumn{1}{l}{}\tabularnewline

 Saponaro \etal \cite{human_intentions_gesture_2013}
& \multicolumn{1}{|c}{\checkmark}  & \multicolumn{1}{c|}{}  & & \multicolumn{1}{c|}{\checkmark } &  \checkmark & & \multicolumn{1}{c|}{ } & \checkmark & &\multicolumn{1}{c|}{}  &\checkmark  & \multicolumn{1}{l}{}\tabularnewline

Stark \etal \cite{Grasping_functionalObectsonlearnedcues_2008}
& \multicolumn{1}{|c}{\checkmark}  & \multicolumn{1}{c|}{}  & & \multicolumn{1}{c|}{\checkmark } &   & & \multicolumn{1}{c|}{\checkmark } & \checkmark & &\multicolumn{1}{c|}{}  &\checkmark  & \multicolumn{1}{l}{}\tabularnewline

\bottomrule
\end{tabular*}}
\caption{Comparison between function-scene understanding methods}
\label{table:functionscenecomparison}
\end{table}

\section{Datasets}\label{Datasets}

In this section, we investigate the available datasets provided with affordance annotations. As the following (Table \ref{table:datasets}) shows,
the distribution of them range from RGB, RGB-D for images and videos. For visual cues, many datasets have been proposed such as UMD and IIT-AFF \cite{AffDetectGeometricFeatures_2015,Object_based_Affordances_CNN_CRF_2017} to facilitate detecting affordance objects from the scene i.e. detecting objects that bear affordances or functionalities from the input image. In other words, these datasets enable researchers to treat affordances as traditional image detection tasks like pedestrian detection or face detection. Since physical and material attributes are important for defining the functionality and affordances, it has been provided in these methods \cite{Visual_Physical_Affordances_2011,ADE20K_Dataset_Object_Parts_2016,See_Glass_Half_Full_2017, What_is_Where_2016}. 
Regarding human activities recognition, these methods \cite{HRI_SocialAffordances_2016, Affordanceto_Improve_Recognition_2011, Weakly_Supervised_Learning_Affordance_2016} advanced annotated datasets that contain human subjects. In the essence of objects parts, this dataset \cite{ADE20K_Dataset_Object_Parts_2016} has part's annotations. 

\begin{table*}
\begin{tabular}{>{\raggedright}m{0.19\textwidth}>{\raggedright}m{0.07\textwidth}>{\raggedright}m{0.17\textwidth}>{\raggedright}m{0.05\textwidth}>{\raggedright}p{0.1\textwidth}>{\raggedright}p{0.03\textwidth}>{\raggedright}p{0.02\textwidth}l>{\raggedright}p{0.07\textwidth}l>{\raggedright}p{0.07\textwidth}}
\toprule
\multicolumn{1}{c|}{\rotatebox[origin=c]{90}{Reference}} 
&\multicolumn{1}{c|}{\rotatebox[origin=c]{90}{Year}} 
&\multicolumn{1}{c|}{\rotatebox[origin=c]{90}{Properties}} & \multicolumn{1}{c|}{\rotatebox[origin=c]{90}{Affordances}} & \multicolumn{1}{c|}{\rotatebox[origin=c]{90}{Format}} & \multicolumn{1}{c|}{\rotatebox[origin=c]{90}{Subjects Number}} & \multicolumn{1}{c|}{\rotatebox[origin=c]{90}{Categories}} & \multicolumn{1}{c|}{\rotatebox[origin=c]{90}{Image/Video}} & \multicolumn{1}{c}{\rotatebox[origin=c]{90}{Subjects}}\tabularnewline
\midrule
 
\\
UMD\cite{AffDetectGeometricFeatures_2015}& 2015 & visual  & Table \ref{table:affordances}  & RGB-D Image  &  & 17 & 30,000 & Indoor\tabularnewline
 
\\
\cite{Visual_Physical_Affordances_2011} &2011 & visual, physical and material & Table \ref{table:affordances} & RGB-D Image  &  & 7 & 375 & Indoor\tabularnewline
 
\\
CAD120 \cite{Weakly_Supervised_Learning_Affordance_2016}&2016 & visual, human-interactions & Table \ref{table:affordances}  & RGB-D Video & 35  &  & 3090 / 215 & Indoor\tabularnewline
 
\\
IIT-AFF \cite{Object_based_Affordances_CNN_CRF_2017} &2017 & visual  & Table \ref{table:affordances}  & RGB-D Image &  & 9 & 8,835 & Tools\tabularnewline
 
\\
ADE- Affordance \cite{ADE20K_Dataset_Object_Parts_2016}& 2016& {visual, physical, social, object-action pairs, exceptions, explanations } & Table \ref{table:affordances}  & RGB Images &  & 7 & 10,000 & Indoors\tabularnewline
 
\\
\cite{Grasping_Local_Global_Features_2016} &2016 & visual &Table \ref{table:affordances}  & RGB Image &  & 8 & 10,360 & Indoor tools\tabularnewline

\\
Extended NYUv2 \cite{multi_scale_affordance_segmentation_2016} & 2016& visual & Table \ref{table:affordances} &  RGB Image &  &5  & 1449 & Indoor\tabularnewline
 
\\
CONTACT VMGdB \cite{Affordanceto_Improve_Recognition_2011} &2011 & visual, human interactions & Table \ref{table:affordances}  & RGB-D Video & 20  & 5 & 5200 & Indoor\tabularnewline
 
\\
HHOI\cite{HRI_SocialAffordances_2016}&2016  & visual, human interactions &Table\ref{table:affordances}  & RGB-D video  & 14  & 5 & -- & Indoor\tabularnewline

\\
Binge Watching \cite{BingeWang_2017_CVPR}& 2017 & visual, human poses & 30 poses  & RGB-D &  & 30 & 11449 & indoor\tabularnewline
 
\\
CERTH-SOR3D \cite{Affordance_Sensorimotor_Recognition_2017} &2017& visual, human interactions & Table \ref{table:affordances}   & RGB-D &  & 14 & 20,800 & indoor, tools\tabularnewline

\\
COQE \cite{See_Glass_Half_Full_2017}&  2017& visual, physical  & --  & RGB &  & 10 & 5000  & Containers\tabularnewline
\\
\cite{What_is_Where_2016} &2016& visual, physical & --  & RGB-D video &  & 4 & 1326 & Containers\tabularnewline
\\
Tool \& Tool-Use (TTU) \cite{Understanding_tools_Affordances_2015}  &2015& visual, physical, human demonstrations  &Table \ref{table:affordances}    & RGB-D Image &  &10  &452   &Tools \tabularnewline
\\

\hline 
\end{tabular}

\caption{The datasets that have ground-truth annotation for affordances and functionalities.}
\label{table:datasets}
\end{table*}

\section{Open Problems and Research Questions}\label{Open Problems}
\textbf{Lack of Consensus on Affordance Definition:} Nearly 5 decades after the introduction of the affordance concept by Gibson, no formal definition of affordances has been agreed upon by AI researchers and ecological psychologists.  Gibson's own description of the concept in his seminal work ``The Ecological Approach to Visual Perception'' (1979) shows the complex nature of the concept, e.g., he said ``Affordance is equally a fact of the environment and a fact of behavior. It is both physical and psychological, yet neither. An affordance points both ways, to the environment and to the observer" \cite{AffordanceIsTargetofCV_1979}. 
Previous efforts have tried to describe affordance as the property of the environment, the mutual phenomenon between an agent and its environment or an observer's perception of the relationships between an agent and its surroundings \cite{chemero2003outline,Formalism_Definition_Sahin_2007}. However, there does not exist a unified definition for affordances so far.  

\textbf{Function vs Affordance:} The terms, function and affordance, are sometimes used with the same meaning in the literature even though they are completely different. The relation between these two terms is complicated where one object may have different affordances and functions such as the case of a hammer which has affordances (grasping, striking, dragging) and functions (drive nails, fit parts, forge metal, and break apart objects). Differently, some objects have affordances and functions with the same meaning such as knife which has affordance (cut) and function (cut). However, the agreed concept (also emphasized in this survey) is that affordance always relate to the object itself rather than function which relates to only another object. In other words, affordance relates to the possible actions whereas the function relates to the effect. Much effort has to be devoted in this issue to distinguish between affordance and functionality in the right manner.

\textbf{Affordance and Attributes:} Recently, much research have been introduced to describe objects with their attributes \cite{ Describing_Objects_Attributes_2009, SUN_Attributes_Dataset_2017,AttributeRecognitionbyJRL2017,HumanAtrrbyHieraContext_2016}. Along with the object label, semantically meaningful attributes are needed to understand the object characteristics. For instance, a cup of tea may be described with some attributes like glass, white and has handle. Therefore, describing the cup by its attributes will help in deciding the best way of grasping. Furthermore, the affordance learning problem needs attributes to address some difficulties such as getting some fruits from the fridge requires prior knowledge about the height of the fridge and the agent as well. Despite its significance, the use of attributes has not been addressed in the context of visual affordances before.

\textbf{Multi-class Labeling:} Considering affordances and functions together gives rise to an advanced set of labels for every object. Therefore, it is a multi-label problem in its nature. For example, the Figure of cup \ref{fig:multifuncti} shows nine labels. If more detailed analysis is required, these labels need to be ordered according to the object, scene and the situation. Remarkably, these labels should be identified in terms of the parts rather than the whole object. 

\textbf{Deep Learning for Affordance Learning:} Nowadays, deep learning is dominating the field of vision and it achieved remarkable improvements in this context. However, it has not received much attention in addressing the challenges particular to affordance and function understanding. By looking at comparison tables, the number of feature learning methods proposed in the literature is too few. Another related factor to deep learning is the size of datasets. In other words, modeling the affordance algorithms with deep learning needs large annotated datasets which are currently unavailable.

\textbf{Complex Affordances:} Affordances of an object can be classified to basic and higher order affordances \cite{AffDetectGeometricFeatures_2015, DFSU2017} e.g. the "pouring" hot water into a cup is basic affordance while making tea from this water is higher-order because it depends on the basic one. These higher-order affordances have not been explored in the existing works. 

 \textbf{Outdoor Affordances:} Mostly all of the previous studies focused on indoor scenes. Only one method has been introduced to address the outdoor affordances \cite{DeepDriving_Affordances_2015}. Despite of less focus on outdoor affordances, this direction of research is highly valuable due to the relationship between affordances and silent actions which is important in self-driving cars, autonomous driving, traffic monitoring applications.
 
\textbf{Affordances for Developmental Robotics:} Since developmental robotics and affordances are tightly related to each other, visually understanding the environment can enhance the learning process. Developmental robotics seek to enhance robot understanding through the environmental interactions while affordances are emergent environmental variables. Thus, learning affordances visually would shorten the time required for interactive robots to build their knowledge accurately. Yet, the conducted studies in this paradigm are still not sufficient and need more investigations to get efficient baselines. In case of extrinsic motivation, visual affordances can be used as reward evaluators and indicators to measure the progress of that learning scheme. For intrinsic motivations, the visual affordances can be utilized to point out the most important features in the environment which will increase the robots curiosity to learn more.

\textbf{Visual Questions Answering (VQA) \& Learning-by-Asking (LBA):} VQA deals with answering intelligent questions about a visual element. Because affordances are constant environment variables that provide highly valuable information about scene content, they can assist in developing a deep insight. Similar to affordance-based recognition, fusing affordances and VQA will improve the accuracy of these answers. Unlike VQA, LBA seeks to understand the environment by asking questions and requesting supervision. Assuming that affordances are a precursor for an object interaction, merging affordances and LBA will reduce the agent time to build its knowledge. 

We believe that the application of visual scene understanding algorithms including those for semantic/instance segmentation, physics based reasoning and 3D volumetric analysis  to affordance will help in resolving several underlying challenges.

\section{Conclusion}\label{Conclusion}
In this survey, visual affordance and functional scene understanding has been reviewed. We introduce a hierarchical taxonomy and cover the progress according to each sub task e.g., classification, segmentation and detection. We begin with formal definition of each sub-task and provide significance and applications to motivate the readers. The paper succinctly compares best approaches in every sub-category in a tabular form to help researchers identify research gaps and open questions. The paper also covers datasets proposed in the field and provides detailed comparisons between them. We discussed the open problems and challenges in this field. Finally, we hope this survey will be helpful for researchers particularly as it is the first survey to review visual affordances and function understanding. 

\ifCLASSOPTIONcaptionsoff
  \newpage
\fi

\bibliographystyle{IEEEtran}
\nocite{*}
\bibliography{DB}

\begin{thebibliography}{100}
\bibitem{What_is_Where_2016}
------, ``What is where: Inferring containment relations from videos.''
\end{thebibliography}

\end{document}